%% file: 3DPAF-arxiv.tex
\pdfoutput=1

\documentclass[10pt,twocolumn,letterpaper]{article}

\usepackage{cvpr}
\usepackage{times}
\usepackage{epsfig}
\usepackage{graphicx}
\usepackage{amsmath}
\usepackage{amssymb}
\usepackage{booktabs} 
\usepackage{amsfonts}
\usepackage{subcaption}
\usepackage{tabularx}
\usepackage{appendix}
\newcolumntype{L}[1]{>{\hsize=#1\hsize\raggedright\arraybackslash}X}%
\newcolumntype{R}[1]{>{\hsize=#1\hsize\raggedleft\arraybackslash}X}%
\newcolumntype{C}[1]{>{\hsize=#1\hsize\centering\arraybackslash}X}%

\usepackage[breaklinks=true,bookmarks=false]{hyperref}

\cvprfinalcopy 

\def\cvprPaperID{****} 

\begin{document}

\title{Monocular Total Capture: Posing Face, Body, and Hands in the Wild}

\author{Donglai Xiang ~ Hanbyul Joo ~ Yaser Sheikh\\
Carnegie Mellon University\\
{\tt\small \{donglaix,hanbyulj,yaser\}@cs.cmu.edu}
}

\maketitle

\begin{abstract}
We present the first method to capture the 3D total motion of a target person from a monocular view input. Given an image or a monocular video, our method reconstructs the motion from body, face, and fingers represented by a 3D deformable mesh model. We use an efficient representation called 3D Part Orientation Fields (POFs), to encode the 3D orientations of all body parts in the common 2D image space. POFs are predicted by a Fully Convolutional Network (FCN), along with the joint confidence maps. To train our network, we collect a new 3D human motion dataset capturing diverse total body motion of 40 subjects in a multiview system.  We leverage a 3D deformable human model to reconstruct total body pose from the CNN outputs by exploiting the pose and shape prior in the model. We also present a texture-based tracking method to obtain temporally coherent motion capture output. We perform thorough quantitative evaluations including comparison with the existing body-specific and hand-specific methods, and performance analysis on camera viewpoint and human pose changes. Finally, we demonstrate the results of our total body motion capture on various challenging in-the-wild videos. Our code and newly collected human motion dataset will be publicly shared.
\end{abstract}

\input{intro}

\input{method}

\input{result}

{\small
\bibliographystyle{ieee}
\bibliography{egbib}
}

\clearpage
\begin{appendices}
\input{supplementary.tex}
\end{appendices}

\end{document}

%% file: intro.tex
\section{Introduction}


Human motion capture is essential for many applications including visual effects, robotics, sports analytics, medical applications, and human social behavior understanding. However, capturing 3D human motion is often costly, requiring a special motion capture system with multiple cameras. For example, the most widely used system \cite{VICON} needs multiple calibrated cameras with reflective markers carefully attached to the subjects' body. The actively-studied markerless approaches are also based on multi-view systems~\cite{Gall-09,Liu-2013,Elhayek-15, joo2017panoptic, joo2018} or depth cameras~\cite{Shotton2011,Baak2011}. For this reason, the amount of available 3D motion data is extremely limited. Capturing 3D human motion from single images or videos can provide a huge breakthrough for many applications by increasing the accessibility of 3D human motion data, especially by converting all human-activity videos on the Internet into a large-scale 3D human motion corpus.

Reconstructing 3D human pose or motion from a monocular image or video, however, is extremely challenging due to the fundamental depth ambiguity. Interestingly, humans are able to almost effortlessly reason about the 3D human body motion from a single view, presumably by leveraging strong prior knowledge about feasible 3D human motions. Inspired by this, several learning-based approaches have been proposed over the last few years to predict 3D human body motion (pose) from a monocular video (image)~\cite{Taylor2000,Ramakrishna2012,akhter2015pose,Tekin2016,Bogo2016,Mehta2017, martinez2017simple,zhou2017towards, kanazawa2018end, Moreno-noguer2017} using available 2D and 3D human pose datasets~\cite{Andriluka-14, lin2014microsoft, CMUMocap, h36m_pami, joo2017panoptic}. Recently, similar approaches have been introduced to predict 3D hand poses from a monocular view~\cite{zimmermann2017learning, mueller2018ganerated, Cai_2018_ECCV}. However, fundamental difficulty still remains due to the lack of available in-the-wild 3D body or hand datasets that provide paired images and 3D pose data; thus most of the previous methods only demonstrate results in controlled lab environments. Importantly, there exists no method that can reconstruct motion from all body parts including body, hands, and face altogether in a single view, although this is important to fully understand human behavior. 

In this paper, we aim to reconstruct the \textbf{3D total motions}~\cite{joo2018} of a human using a monocular imagery captured in the wild. This ambitious goal requires solving challenging 3D pose estimation problems for different body parts altogether, which are often considered as separate research domains. Notably, we apply our method to in-the-wild situations (e.g., videos from YouTube), which has rarely been demonstrated in previous work. We use a 3D representation named Part Orientation Fields (POFs) to efficiently encode the 3D orientation of a body part in the 2D space. A POF is defined for each body part that connects adjacent joints in torso, limbs, and fingers, and represents relative 3D orientation of the rigid part regardless of the origin of 3D Cartesian coordinates. POFs are efficiently predicted by a Fully Convolutional Network (FCN), along with 2D joint confidence maps~\cite{tompson2014joint, Wei2016, cao2017realtime}. To train our networks, we collect a new 3D human motion dataset containing diverse body, hands, and face motions from 40 subjects. Separate CNNs are adopted for body, hand and face, and their outputs are consolidated together in a unified optimization framework. We leverage a 3D deformable model that is built for the total capture~\cite{joo2017panoptic} in order to exploit the shape and motion prior embedded in the model. In our optimization framework, we fit the model to the CNN measurements at each frame to simultaneously estimate the 3D motion of body, face, fingers, and feet. Our mesh output also enables us to additionally refine our motion capture results for better temporal coherency by optimizing the photometric consistency in the texture space.

This paper presents the first approach to monocular total motion capture in various challenging in-the-wild scenarios (e.g., Fig.~\ref{fig:teaser}). We demonstrate that our single framework achieves comparable results to existing state-of-the-art 3D body or hand pose estimation methods on public benchmarks. Notably, our method is applied to various in-the-wild videos, which has rarely been demonstrated in either 3D body or hand estimation area. We also conduct thorough experiments on our newly collected dataset to quantitatively evaluate the performance of our method with respect to viewpoint and body pose changes. The major contributions of our paper are summarized as follows:

\begin{itemize}
\item We present the first method to produce \textbf{3D total motion capture} results from a monocular image or a video in various challenging in-the-wild scenarios.
\item We introduce an optimization framework to fit a deformable human model on 3D POFs and 2D keypoint measurements for total body pose estimation, and show comparable results to the state-of-the-art methods in both 3D body and 3D hand estimation benchmarks.
\item We present a method to enforce photometric consistency across time to reduce motion jitters.
\item We capture a new 3D human motion dataset with 40 subjects to provide training and evaluation data for monocular total motion capture. 
\end{itemize}

\section{Related Work}

\textbf{Single Image 2D Human Pose Estimation:} Over the last few years, great progress has been made in detecting 2D human body keypoints from a single image~\cite{toshev2014deeppose,tompson2014joint,bulat2016human,Wei2016,Newell-16,cao2017realtime} by leveraging large-scale manually annotated datasets~\cite{lin2014microsoft,Andriluka-14} with deep Convolutional Neural Network (CNN) framework. In particular, the major breakthrough is boosted by using the fully convolutional architectures to produce confidence scores for each joint with a heatmap representation~\cite{tompson2014joint,Wei2016,Newell-16,cao2017realtime}, which is known to be more efficient than directly regressing the joint locations with fully connected layers~\cite{toshev2014deeppose}. A recent work~\cite{cao2017realtime} similarly learns the connectivity between pairs of adjacent joints, called the Part Affinity Fields (PAFs) in the form of 2D heatmaps, to assemble 2D keypoints for different individuals in the multi-person 2D pose estimation problem.


\begin{figure*}[t]
 \centering
   \includegraphics[width=\linewidth]{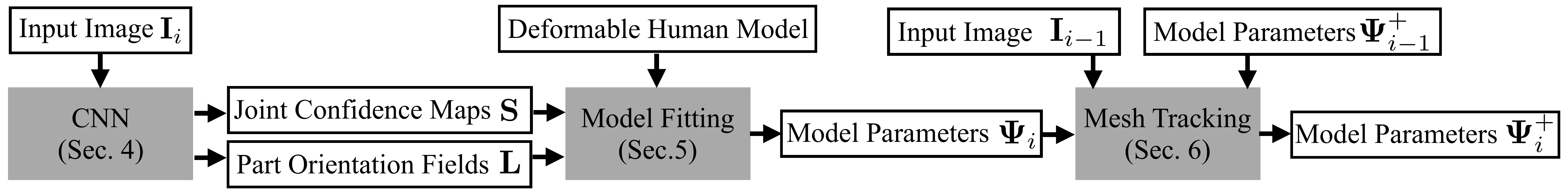}
   \caption{An overview of our method. Our method is composed of CNN part, mesh fitting part, and mesh tracking part.} 
   \label{fig:overview}
\end{figure*}

\textbf{Single Image 3D Human Pose Estimation:} Early work~\cite{Ramakrishna2012,akhter2015pose} models the 3D human pose space as an over-complete dictionary learned from a 3D human motion database~\cite{CMUMocap}. More recent approaches rely on deep neural networks, which are roughly divided into two directions: two-stage methods and direct estimation. The two-stage methods take 2D keypoint estimation as input and focus on lifting 2D human poses to 3D independently without input image \cite{Bogo2016,Chen2017,martinez2017simple,Moreno-noguer2017,Nie2017,AAAI1816471}. These methods ignore rich information in images that encodes 3D information, such as shading and appearance, and also suffer from sensitivity to 2D localization error. Direct estimation methods predict 3D human pose directly from images, in the form of direct coordinate regression \cite{Rogez2016, Sun2017, Sun_2018_ECCV}, voxel prediction \cite{pavlakos2017coarse, luvizon20182d, varol2018bodynet} or depth map prediction \cite{zhou2017towards}. Similar to ours, a recent work uses 3D orientation fields~\cite{orinet2018} as an intermediate representation for the 3D body pose. However, these models are usually trained on MoCap datasets, with limited ability to generalize to in-the-wild scenarios.

Due to the above limitations, some methods have been proposed to integrate prior knowledge about human pose for better in-the-wild performance. Some work \cite{pavlakos2018ordinal, relativeposeBMVC18, ijcai2018-136} proposes to use ordinal depth as additional supervision for CNN training. Additional loss functions are introduced in \cite{zhou2017towards, dabral2018learning} to enforce constraints on predicted bone length and joint angles. Some work \cite{kanazawa2018end, yang20183d} uses Generative Adversarial Networks (GAN) to exploit human pose prior in data-driven approaches.


\textbf{Monocular Hand Pose Estimation:} Hand keypoint estimation is often considered as independent research domain from body pose estimation. Most of previous work is based on depth image as input~\cite{Oikonomidis-12,Sridhar-13,Sharp-15,Sridha-15,Tzionas-16,Ye-16}, while RGB-based method is introduced recently, for 2D keypoint estimation~\cite{simon2017hand} and 3D pose estimation~\cite{zimmermann2017learning, Cai_2018_ECCV, Iqbal_2018_ECCV}.

\textbf{3D Deformable Human Models:} 3D deformable models are commonly used for markerless body~\cite{anguelov2005scape, Loper2015, pons2015dyna} and face motion capture~\cite{blanz1999morphable, cao2014facewarehouse} to restrict the reconstruction output to the parametric shape and motion spaces defined by the models. Although the outputs are limited by the expressive power of models (e.g., some body models cannot express clothing and some face models cannot express wrinkles), they greatly simplify the 3D motion capture problem. We can fit the models based on available measurements by optimizing cost functions with respect to the model parameters. Recently, a generative 3D model that can express body and hands is introduced by Romero et al.~\cite{romero2017embodied}; the Adam model is introduced by Joo et al.~\cite{joo2018} to enable the total body motion capture (face, body and hands), which we adopt for monocular total capture.

%% file: method.tex
\section{Method Overview}
Our method takes as input a sequence of images capturing the motion of a single person from a monocular RGB camera, and outputs the 3D total body motion capture (including the motion from body, face, hands, and feet) of the target person in the form of a deformable 3D human model~\cite{Loper2015,joo2018} for each frame. Given a $N$-frame video sequence, our method produces the parameters of the 3D human body model~\cite{joo2018}, including body motion parameters $\{\boldsymbol{\theta}_i\}_{i=1}^N$, facial expression parameters $\{\boldsymbol{\sigma}_i\}_{i=1}^N$, and global translation parameters $\{\boldsymbol{t}_i\}_{i=1}^N$. The body motion parameter $\boldsymbol{\theta}$ includes hands and feet motions, as well as the global rotation of the body. Our method also estimates shape coefficients $\boldsymbol{\phi}$ shared among all frames in the sequence, while $\boldsymbol{\theta}$, $\boldsymbol{\sigma}$, and $\boldsymbol{t}$ are estimated for each frame respectively. The output parameters are defined by the 3D deformable human model Adam \cite{joo2018}. Note that our method can be also applied to capture only a subset of total motions (e.g., body motion only with the SMPL model~\cite{Loper2015} or hand motion only by separate hand model of Frankenstein in \cite{joo2018}). We denote a set of all parameters $(\boldsymbol{\phi}, \boldsymbol{\theta}, \boldsymbol{\sigma}, \boldsymbol{t})$ by $\boldsymbol{\Psi}$, and denote the result for the $i$-th frame by $\boldsymbol{\Psi}_i$.

Our method is divided into 3 stages, as shown in Fig.~\ref{fig:overview}. In the first stage, each image is fed into a Convolutional Neural Network (CNN) obtain the joint confidence maps and the 3D orientation information of body parts, which we call the 3D Part Orientation Fields (POFs). In the second stage, we perform total body motion capture by fitting a deformable human mesh model ~\cite{joo2018} on the image measurements produced by the CNNs. We utilize the prior information embedded in the human body model for better robustness of results against the noise in CNN outputs. This stage produces the 3D pose for each frame independently, represented by parameters of the deformable model $\{\boldsymbol{\Psi}_i\}_{i=1}^N$. In the third stage, we additionally enforce temporal consistency across frames to reduce motion jitters. We define a cost function to ensure photometric consistency in the texture domain of mesh model, based on the initial fitting outputs of the second stage. This stage produces refined model parameters $\{\boldsymbol{\Psi}^+_i\}_{i=1}^N$. We demonstrate that this temporal refinement is crucial to obtain realistic body motion capture output.

\section{Predicting 3D Part Orientation Fields}
\label{sec:3DBOF}

\begin{figure}[t]
   \includegraphics[width=\columnwidth]{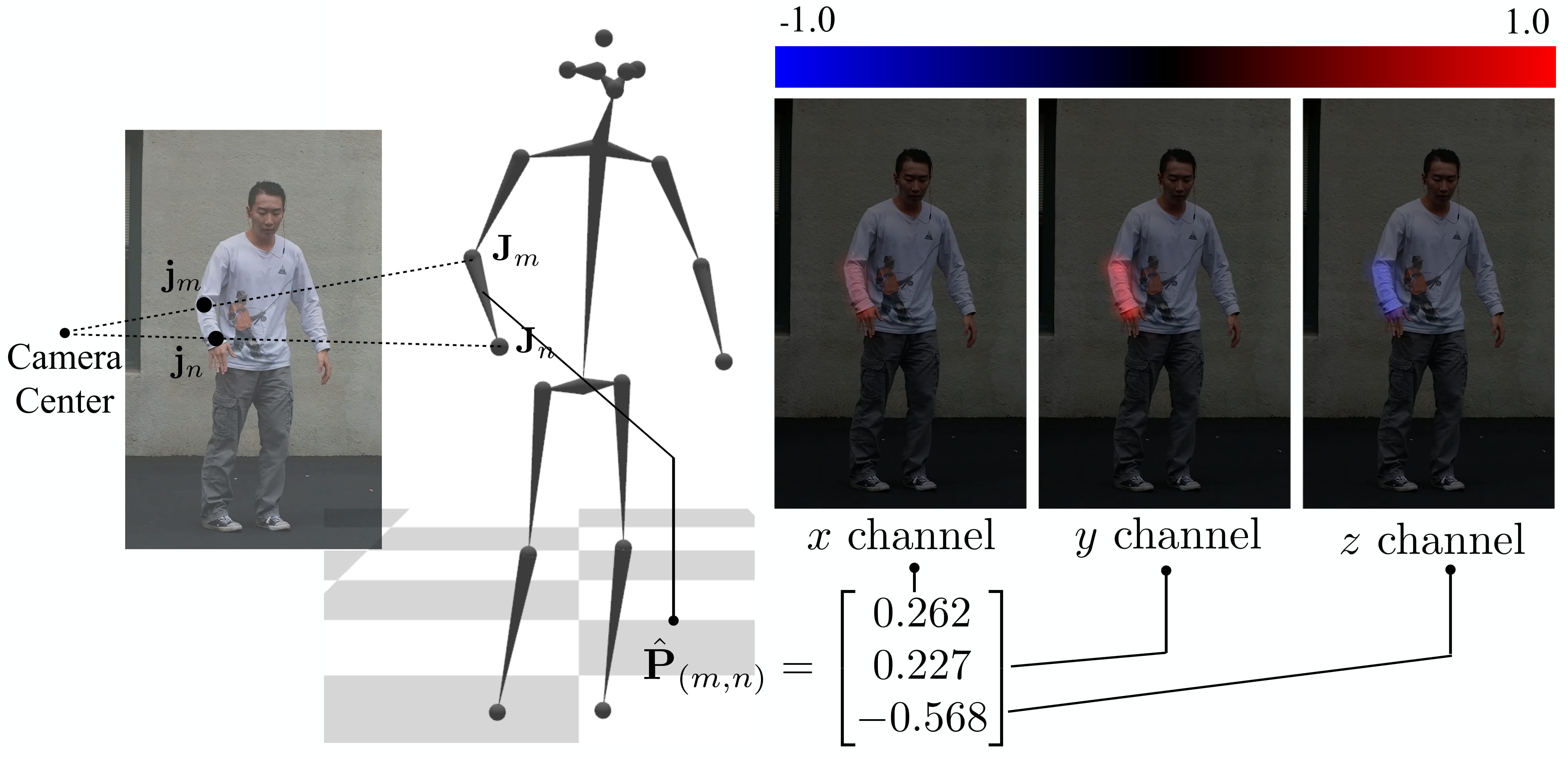}
   \caption{An illustration of a Part Orientation Field. The orientation $\hat{\mathbf P}_{(m,n)}$ for body part $\mathbf P_{(m,n)}$ is a unit vector from $\mathbf J_m$ to $\mathbf J_n$. In POFs, all pixels belong to this part are assigned the value of this vector in $x, y, z$ channels.}
   \label{fig:3dbof}
\end{figure}


The 3D Part Orientation Field (POF) encodes the 3D orientation of a body part of an articulated structure (e.g., limbs, torso, and fingers) in 2D image space. The same representation is used in a very recent literature~\cite{orinet2018}, and we describe the details and notations used in our total motion capture framework. We pre-define a human skeleton hierarchy $\mathbb S$ in the form of a set of `(parent, child)' pairs\footnote{See the appendix for our body and hand skeleton definition.}. A rigid body part connecting a 3D parent joint $\mathbf J_m \in \mathbb R^3$ and a child joint $\mathbf J_n \in \mathbb R^3$, $(m, n) \in \mathbb S$, is denoted by $\mathbf P_{(m,n)}$, with $\mathbf J_m, \mathbf J_n$ defined in the camera coordinate. Its 3D orientation $\hat{\mathbf{P}}_{(m,n)}$ is represented by a unit vector from $\mathbf J_m$ to $\mathbf J_n$ in $\mathbb R^3$ :
\begin{align}
\hat{\mathbf{P}}_{(m,n)} = \frac{\mathbf{J}_n - \mathbf{J}_m}{||\mathbf{J}_n - \mathbf{J}_m||}~.
\end{align}
For a specific body part $\mathbf P_{(m,n)}$, we define a Part Orientation Field $\mathbf L_{(m,n)} \in \mathbb{R}^{3 \times h \times w}$ to represent its 3D orientation $\hat{\mathbf{P}}_{(m,n)}$ as a 3-channel heatmap (for $x, y, z$ coordinates respectively) in the image space, where $h$ and $w$ are the size of image. The value of the heatmap at $\mathbf x$ in the POF $\mathbf L_{(m,n)}$ is defined as,
\begin{equation}
\mathbf{L}_{(m,n)}(\mathbf{x}) = 
\begin{cases}
\hat{\mathbf{P}}_{(m,n)} & \text{ if } \mathbf{x} \in \mathbf{P}_{(m,n)},\\
\mathbf 0 & \text{otherwise.}
\end{cases}
\end{equation}
Note that the POF values are defined only for the pixel region belonging to the current target part $\mathbf P_{(m,n)}$ and we follow \cite{cao2017realtime} to define the pixels belonging to the part as a rectangular (please refer to \cite{cao2017realtime} for details).
An example POF of a body part (right lower arm) is shown in Fig. \ref{fig:3dbof}.

\noindent \textbf{Implementation Details:} We train a CNN to predict joint confidence maps $\mathbf S$ and Part Orientation Fields $\mathbf L$. The input image is cropped around the target person to $368 \times 368$, with the bounding box given by OpenPose\footnote{\url{https://github.com/CMU-Perceptual-Computing-Lab/openpose}} \cite{cao2017realtime, simon2017hand} during testing. We follow \cite{cao2017realtime} for CNN architecture with minimum change. We use 3 channels to estimate POF instead of 2 channels in \cite{cao2017realtime} for every body part in $\mathbb S$. $L_2$ loss is applied to network prediction on $\mathbf S$ and $\mathbf L$. We also train on our network on images with 2D pose annotations (e.g. COCO). In this situation we only supervise the network with loss on $\mathbf S$. Two networks are trained for body and hands separately.

\section{Model-Based 3D Pose Estimation}
\label{sec:Fit}
Ideally the joint confidence maps $\mathbf{S}$ and POFs $\mathbf{L}$ produced by CNN provide sufficient information to reconstruct a 3D skeletal structure up to scale~\cite{orinet2018}. In practice, the $\mathbf{S}$ and $\mathbf{L}$ can be noisy, so we exploit a 3D deformable mesh model to more robustly estimate 3D human pose with the shape and pose priors embedded in the model. In this section, we first describe our mesh fitting process for body, and then extend it to hand pose and facial expression for total body motion capture. We use superscripts $B, LH, RH, T$ and $F$ to denote functions and parameters for body, left hand, right hand, toes, and face respectively. We use Adam \cite{joo2018} which encompasses the expressive power for body, hands and facial expression in a single model. Other human models (e.g., SMPL~\cite{Loper2015}) can be also used if the goal is to reconstruct only part of the total body motion.


\subsection{Deformable Mesh Model Fitting with POFs}
\begin{figure}[t]
   \includegraphics[width=\columnwidth]{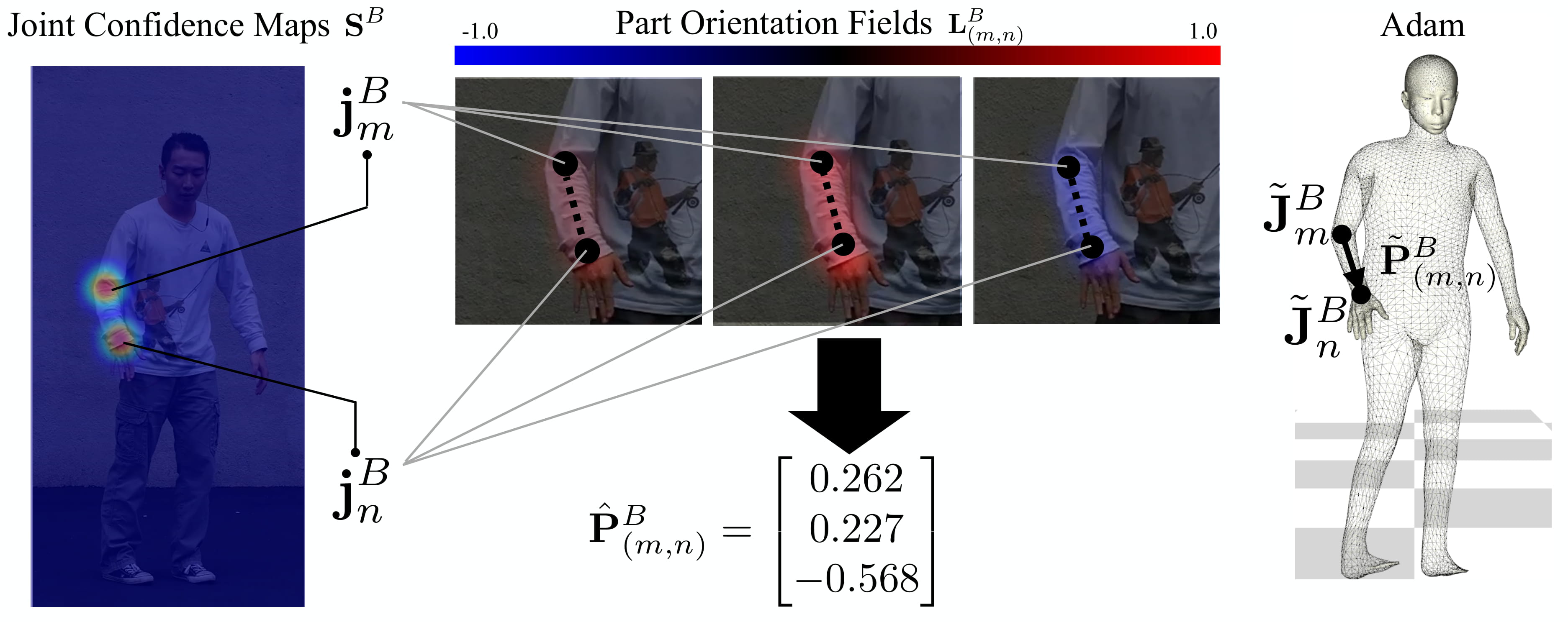}
   \caption{Human model fitting on estimated POFs and joint confidence maps. We extract 2D joint locations from 2D joint confidence maps (left) and then body part orientation from POFs (right). Then we optimize a cost function (Equation \ref{eq:modelfitting}) that minimizes the distance between $\mathbf{\Pi}(\tilde{\mathbf{J}}^B_m)$ and $\mathbf{j}^B_m$ and angle between $\tilde{\mathbf{P}}^B_{(m,n)}$ and $\hat{\mathbf{P}}^B_{(m,n)}$.}
   \label{fig:modelfitting}
\end{figure}

Given the 2D joint confidence maps $\mathbf S^B$ predicted by our CNN for body, we obtain 2D keypoint locations $\{\mathbf j^B_m\}_{m=1}^J$ by taking the maximum in each channel of $\mathbf S^B$. Given the $\{\mathbf j^B_m\}_{m=1}^J$ and the other CNN output POFs $\mathbf L^B$, we compute the 3D orientation of each bone ${\hat{\mathbf P}}^B_{(m,n)}$, by averaging the values of $\mathbf L^B$ along the segment from $\mathbf j^B_m$ to $\mathbf j^B_n$, as in \cite{cao2017realtime}. We obtain a set of mesh parameters $\boldsymbol{\theta}$, $\boldsymbol{\phi}$, and $\boldsymbol{t}$ that agree with these image measurements by minimizing the following objective function:
\begin{equation}
\begin{aligned}
\mathcal{F}^B(\boldsymbol{\theta}, \boldsymbol{\phi}, \boldsymbol{t}) = \mathcal{F}^B_{\text{2D}}(\boldsymbol{\theta}, \boldsymbol{\phi}, \boldsymbol{t}) +  \mathcal{F}^B_{\text{\tiny POF}}(\boldsymbol{\theta}, \boldsymbol{\phi}) +  \mathcal{F}^B_p(\boldsymbol{\theta}),
\label{eq:modelfitting}
\end{aligned}
\end{equation}
where $\mathcal{F}^B_{\text{2D}}$, $\mathcal{F}^B_{\text{\tiny POF}}$, and $\mathcal{F}^B_{p}$ are different constraints as defined below. The 2D keypoint constraint $\mathcal{F}^B_{\text{2D}}$ penalizes the discrepancy between network-predicted 2D keypoints and the projections of the joints in the human body model:
\begin{align}
\mathcal{F}^B_{\text{2D}}(\boldsymbol{\theta}, \boldsymbol{\phi}, \boldsymbol{t}) = \sum_m \Vert\mathbf{j}^B_m - \mathbf{\Pi}(\tilde{\mathbf{J}}^B_m (\boldsymbol{\theta}, \boldsymbol{\phi}, \boldsymbol{t}))\Vert^2,
\label{eq:modelfitting_2D}
\end{align}
where $\tilde{\mathbf{J}}^B_m (\boldsymbol{\theta}, \boldsymbol{\phi}, \boldsymbol{t})$ is $m$-th joint of the human model and $\mathbf{\Pi}(\cdot)$ is projection function from 3D space to image, where we assume a weak perspective camera model. The POF constraint $\mathcal{F}^B_{\text{\tiny POF}}$ penalizes the difference between POF prediction and the direction of body part in mesh model, defined as:
\begin{align}
\mathcal{F}^B_{\text{POF}}(\boldsymbol{\theta}, \boldsymbol{\phi}) = w^B_{\text{\tiny POF}}\sum_{(m,n) \in \mathbb{S}}  1 - \hat{\mathbf{P}}^B_{(m,n)} \cdot \tilde {\mathbf{P}}^B_{(m,n)} (\boldsymbol{\theta}, \boldsymbol{\phi}),
\label{eq:modelfitting_BOF}
\end{align}
where $\tilde{\mathbf{P}}^B_{(m,n)}$ is the unit directional vector for the bone $\mathbf P^B_{(m,n)}$ in the human mesh model, $w^B_{\text{\tiny POF}}$ is a balancing weight for this term, and $\cdot$ is inner product between 3-vectors. The prior term $\mathcal{F}^B_{p}$ is needed to restrict our output to a feasible human pose distribution (especially for rotation around bones), defined as:
\begin{align}
\mathcal{F}^B_p(\boldsymbol{\theta}) = w^B_{p}\Vert \mathbf A^B_\theta (\boldsymbol\theta - \boldsymbol \mu^B_\theta) \Vert^2,
\label{eq:modelfitting_prior}
\end{align}
where $\mathbf A^B_\theta$ and $\boldsymbol \mu^B_\theta$ are poses prior learned from CMU Mocap dataset~\cite{CMUMocap}, and $w^B_{p}$ is a balancing weight. We use Levenberg-Marquardt algorithm \cite{ceres-solver} to optimize Equation \ref{eq:modelfitting}. The mesh fitting process is illustrated in Fig. \ref{fig:modelfitting}.

\subsection{Total Body Capture with Hands, Feet and Face}


Given the network output of the hand network $\mathbf S^{LH}, \mathbf L^{LH}$ and $\mathbf S^{RH}, \mathbf L^{RH}$ for both hands, we can additionally fit the Adam model to satisfy the hand pose using similar optimization objectives:
\begin{equation}
\begin{aligned}
\mathcal{F}^{LH}(\boldsymbol{\theta}, \boldsymbol{\phi}, \boldsymbol{t}) = \mathcal{F}^{LH}_{\text{2D}}(\boldsymbol{\theta}, \boldsymbol{\phi}, \boldsymbol{t}) + \mathcal{F}^{LH}_{\text{\tiny POF}}(\boldsymbol{\theta}, \boldsymbol{\phi}) +  \mathcal{F}^{LH}_p(\boldsymbol{\theta}).
\end{aligned}
\end{equation}
$\mathcal{F}^{LH}$ is the objective function for left hand and each term is defined similarly to Equation \ref{eq:modelfitting_2D}, \ref{eq:modelfitting_BOF} and \ref{eq:modelfitting_prior}. The hand pose priors are learned from MANO dataset \cite{romero2017embodied}. The objective function for the right hand $\mathcal{F}^{RH}$ is similarly defined. 

Once we fit the body and hand parts of deformable model to the CNN outputs, the projection of the 3D model on the image is already well aligned to the target person. Then we can reconstruct other body parts by simply adding more 2D joint constraints using additional 2D keypoint measurements. In particular, we include 2D face and foot keypoints from the OpenPose detector. The additional cost function for toes is defined as:
\begin{align}
\mathcal{F}^T(\boldsymbol{\theta}, \boldsymbol{\phi}, \boldsymbol{t}) = \sum_m \Vert \mathbf{j}^T_m - \mathbf{\Pi}(\tilde{\mathbf{J}}^T_m (\boldsymbol{\theta}, \boldsymbol{\phi}, \boldsymbol{t})) \Vert^2,
\end{align}
where $\{\mathbf{j}^T_m\}$ are 2D tiptoe keypoints on both feet from OpenPose, and $\{\tilde{\mathbf{J}}^T_m\}$ are the 3D joint location of the mesh model in use. Similarly for face we define:
\begin{align}
\mathcal{F}^F(\boldsymbol{\theta}, \boldsymbol{\phi}, \boldsymbol{t}, \boldsymbol{\sigma}) = \sum_m \Vert \mathbf{j}^F_m - \mathbf{\Pi}(\tilde{\mathbf{J}}^F_m (\boldsymbol{\theta}, \boldsymbol{\phi}, \boldsymbol{t}, \boldsymbol{\sigma})) \Vert^2.
\end{align}
Note that the facial keypoints $\tilde{\mathbf{J}}^F_m$ are determined by all the mesh parameters $\boldsymbol{\theta}, \boldsymbol{\phi}, \boldsymbol{t}, \boldsymbol{\sigma}$ together. In addition, we also apply regularization for shape coefficients and facial expression coefficients:
\begin{align}
R^{\boldsymbol{\phi}}(\boldsymbol{\phi}) = \Vert \boldsymbol{\phi} \Vert^2,
R^{\boldsymbol{\sigma}}(\boldsymbol{\sigma}) = \Vert \boldsymbol{\sigma} \Vert^2.
\end{align}

Putting everything together, the final optimization objective is
\begin{equation}
    \begin{gathered}
        \mathcal{F}(\boldsymbol{\theta}, \boldsymbol{\phi}, \boldsymbol{t}, \boldsymbol{\sigma}) = \mathcal{F}^B +  \mathcal{F}^{LH}+
        \mathcal{F}^{RH} + \\ \mathcal{F}^T +\mathcal{F}^F+
        R^{\boldsymbol{\phi}} + R^{\boldsymbol{\sigma}},
        \label{eq:full}
    \end{gathered}
\end{equation}
where the balancing weights for all the terms are omitted for clarity. We optimize this final objective function in multiple stages to avoid local minima. We first fit the torso, then add limbs, and finally optimize the full objective function including all constraints. This stage produces 3D total body motion capture results for each frame independently in the form of Adam model parameters $\{\mathbf \Psi_i\}_{i=1}^N$.

\section{Enforcing Photo-Consistency in Textures}

In the previous stages, we perform per-frame processing, which is vulnerable to motion jitters. We propose to reduce the jitters using the pixel-level image cues given the initial model fitting results. The core idea is to enforce photometric consistency in textures of the model, extracted by projecting the fitted mesh models on the input images. Ideally, the textures should be consistent across frames, but in practice there exist discrepancies due to motion jitters. In order to efficiently implement this constraint in our optimization framework, we compute optical flows from projected texture to the target input image. The destination of each flow indicates the expected location of vertex projection. To describe our method, we define a function $\mathcal{T}$ which extracts a texture given an image and a mesh structure:
\begin{equation}
\begin{gathered}
\boldsymbol{\mathcal{T}}_i = \mathcal{T} \left( \mathbf{I}_i, M(\boldsymbol{\Psi}_{i}) \right).
\label{eq:texture}
\end{gathered}
\end{equation}
Given the input image $\mathbf I_i$ of the $i$-th frame and mesh determined by $\mathbf \Psi_i$, the function $\mathcal{T}$ extracts a texture map $\boldsymbol{\mathcal{T}}_i$ by projecting the mesh structure for $i$-th frame on the image for the visible parts. We ideally expect the texture for ($i$+$1$)-th frame $\boldsymbol{\mathcal{T}}_{i+1}$ to be the same as $\boldsymbol{\mathcal{T}}_i$. Instead of directly using this constraint for optimization, we use optical flow to compute the discrepancy between these textures for easier optimization. More specifically, we pre-compute the optical flow between the raw image $\mathbf{I}_{i+1}$ and the rendering of the mesh model at ($i$+$1$)-th frame with the $i$-th frame's texture map $\boldsymbol{\mathcal{T}}_i$, which we call `synthetic image':
\begin{equation}
\begin{gathered}
\mathbf{f}_{i+1} = f( \mathcal{R}( M_{i+1}, \boldsymbol{\mathcal{T}}_i), \mathbf{I}_{i+1} ),
\end{gathered}
\end{equation}
where $M_{i+1} = M(\mathbf \Psi_{i+1})$ is the mesh for the ($i$+$1$)-th frame, and $\mathcal{R}$ is a rendering function that renders a mesh with a texture to an image. The function $f$ computes optical flows from the synthetic image to the input image $\mathbf{I}_{i+1}$. The output flow $\mathbf{f}_{i+1}: \mathbf{x} \longrightarrow \mathbf{x'}$ maps a 2D location $\mathbf{x}$ to a new location $\mathbf{x'}$ following the optical flow result. Intuitively, the computed flow mapping $\mathbf{f}_{i+1}$ drives the projection of 3D mesh vertices toward the directions for better photometric consistency in textures across frames. Based on this flow mapping, we define the texture consistency term:
\begin{equation}
\begin{gathered}
\mathcal{F}_{\text{tex}}(\boldsymbol{\Psi}^+_{i+1}) = \sum_{n} \Vert \mathbf v^+_n(i+1) - \mathbf v'_n(i+1)\Vert^2,
\end{gathered}
\label{eq:tracking}
\end{equation}
where $\mathbf v^+_n(i+1)$ is the projection of the $n$-th mesh vertex as a function of model parameters $\boldsymbol{\Psi}^+_{i+1}$ under optimization. $\mathbf v'_n(i+1) = \mathbf f_{i+1}(\mathbf{v}_n(i+1))$ is the destination of each optical flow, where $\mathbf{v}_n(i+1)$ is the projection of $n$-th mesh vertex of mesh $M_{i+1}$. Note that $\mathbf v'_n(i+1)$ is pre-computed and constant during the optimization. This constraint is defined in image space, and thus it mainly reduces the jitters in $x, y$ directions. Since there is no image clue to reduce the jitters along $z$ direction, we just enforce a smoothness constraint for $z$-components of 3D joint locations:
\begin{equation}
\begin{gathered}
    \mathcal{F}_{\Delta z}(\boldsymbol{\theta}^+_{i+1}, \boldsymbol{\phi}^+_{i+1}, \boldsymbol{t}^+_{i+1}) = \sum_m (\mathbf J^{+z}_m(i+1) - \mathbf J^z_m(i))^2,
\end{gathered}
\label{eq:zsmoothing}
\end{equation}
where $\mathbf J^{+z}_m(i+1)$ is $z$-coordinate of the $m$-th joint of the mesh model as a function of parameters under optimization, and $\mathbf J^{z}_m(i)$ is the corresponding value in previous frame as a fixed constant. Finally, we define a new objective function:
\begin{equation}
\begin{gathered}
\mathcal{F}^+ (\boldsymbol{\Psi}^+_{i+1}) = 
\mathcal{F}_{\text{tex}} + \mathcal{F}_{\Delta z} + \mathcal{F}_{\tiny \text{POF}} + \mathcal{F}^F,
\end{gathered}
\label{eq:tracking_total}
\end{equation}
where the balancing weights are omitted. We minimize this function to obtain the parameter of the ($i$+1)-th frame $\mathbf \Psi^+_{i+1}$, initialized from output of last stage $\mathbf \Psi_{i+1}$. Compared to the original full objective Equation ~\ref{eq:full}, this new objective function is simpler since this optimization starts from a good initialization. Most of the 2D joint constraints are replaced by $\mathcal{F}_{\text{tex}}$, while we found that the POF term and face keypoint term are still needed to avoid error accumulation. Note that this optimization is performed recursively---we use the updated parameters of the $i$-th frame $\boldsymbol\Psi^+_i$ to extract the texture $\boldsymbol{\mathcal T}_i$ in Equation \ref{eq:texture}, and update the model parameters at the ($i$+1)-th frame from $\boldsymbol\Psi_{i+1}$ to $\boldsymbol\Psi^+_{i+1}$ with this optimization. Also note that the shape parameters $\{\boldsymbol\phi^+_i\}$ should be the same across the sequence, so we take $\boldsymbol\phi^+_{i+1} = \boldsymbol\phi^+_i$ and fix it during optimization. We also freeze facial expression and does not optimize it in this stage.

\begin{figure}[t]
  \includegraphics[width=\columnwidth]{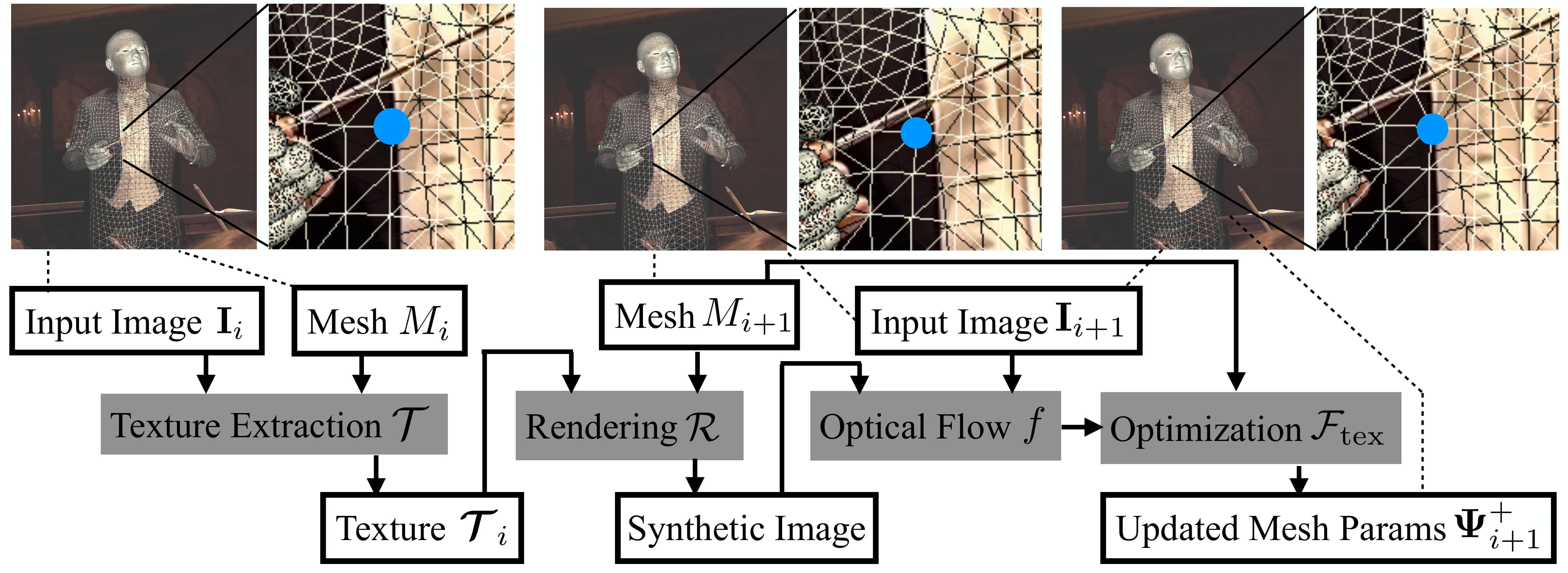}
  \caption{Illustration of our temporal refinement algorithm. The top row shows meshes projected on input images at previous frame, current target frame, and after refinement. In zoom-in views a particular vertex is shown in blue, which is more consistent after applying our tracking method.}
  \label{fig:tracking}
\end{figure}

%% file: result.tex
\section{Results}
We quantitatively evaluate the performance of our method on public benchmarks for 3D body pose estimation and hand pose estimation. We also thoroughly evaluate our method on view point changes and human pose changes in our newly collected multi-view human pose dataset. For all quantitative experiments, we use the camera intrincics provided by the datasets. We finally show our total motion capture results in various challenging videos recorded by us or obtained from YouTube. Our qualitative results are best shown in our supplementary videos.

\subsection{Dataset}

\noindent \textbf{Body Pose Dataset:} \textit{Human3.6M} \cite{h36m_pami} is a large-scale indoor marker-based human MoCap dataset, and currently the most commonly used benchmark for 3D body pose estimation. We quantitatively evaluate the body part of our algorithm on it. We follow the standard training-testing protocol as in \cite{pavlakos2017coarse}.


\noindent \textbf{Hand Pose Dataset:} \textit{Stereo Hand Pose Tracking Benchmark (STB)} 
\cite{zhang20163d} is a 3D hand pose dataset consisting of 30K images for training and 6K images for testing.
\textit{Dexter+Object (D+O)} \cite{sridhar2016real} is a hand pose dataset captured by an RGB-D camera, providing about 3K testing images in 6 sequences. Only the locations of finger tips are annotated.

\noindent \textbf{Newly Captured Total Motion Dataset:} We use the Panoptic Studio \cite{joo2015panoptic, joo2017panoptic} to capture a new dataset for 3D body and hand pose in a markerless way~\cite{joo2018}. We use 31 HD cameras to capture 40 subjects. Each subject makes a wide range of motion in body and hand under the guidance of a video for 2.5 minutes. After cleaning out the erroneous frames, we obtain about 834K body images and 111K hand images with corresponding 3D pose data. We split this dataset into training and testing set such that no subject appears in both. This dataset will be publicly shared.

\begin{table*}[t]
\centering
\scriptsize
\begin{tabularx}{\textwidth}{C{1}|C{1}|C{1}|C{0.8}|C{1}|C{0.8}|C{0.8}|C{1}|C{0.8}|C{0.8}||C{1.4}|C{1}|C{1}|C{0.8}|C{0.8}}
    \hline
    Method & Pavlakos \cite{pavlakos2017coarse} & Zhou \cite{zhou2017towards} & Luo \cite{orinet2018} & Martinez \cite{martinez2017simple} & Fang \cite{AAAI1816471} & Yang \cite{yang20183d} & Pavlakos \cite{pavlakos2018ordinal} & Dabral \cite{dabral2018learning} & Sun \cite{Sun_2018_ECCV} & *Kanazawa \cite{kanazawa2018end} & *Metah \cite{Mehta2017} & *Metah \cite{singleshotmultiperson2018} & *Ours & *Ours+ \\
    \hline
    MPJPE & 71.9 & 64.9 & 63.7 & 62.9 & 60.4 & 58.6 & 56.2 & 55.5 & \textbf{49.6} & 88.0 & 80.5 & 69.9 & \textbf{58.3} & 64.5 \\
    \hline
\end{tabularx}
\caption{Quantitative comparison with previous work on Human3.6M dataset. The `*' signs indicate methods that show results on in-the-wild videos. The evaluation metric is Mean Per Joint Position Error (MPJPE) in millimeter. The numbers are taken from original papers. `Ours' and `Ours+' refer to our results without and with prior respectively.}
\label{table:human3.6m}
\end{table*}

\subsection{Quantitative Comparison with Previous Work}

\subsubsection{3D Body Pose Estimation.}
\textbf{Comparison on Human3.6M.} We compare the performance of our single-frame body pose estimation method with previous state-of-the-arts. Our network is initialized from the 2D body pose estimation network of OpenPose. We train the network using COCO dataset~\cite{lin2014microsoft}, our new 3D body pose dataset, and Human3.6M for 165k iterations with a batch size of 4. During testing time, we fit Adam model~\cite{joo2018} onto the network output. Since Human3.6M has a different joint definition from Adam model, we build a linear regressor to map Adam mesh vertices to 17 joints in Human3.6M definition using the training set, as in \cite{kanazawa2018end}. For evaluation, we follow \cite{pavlakos2017coarse} to rescale our output to match the size of an average skeleton computed from the training set. The Mean Per Joint Position Error (MPJPE) after aligning the root joint is reported as in \cite{pavlakos2017coarse}.

The experimental results are shown in Table \ref{table:human3.6m}. Our method achieves competitive performance; in particular, we show the lowest pose estimation error among all methods that demonstrate their results on in-the-wild videos (marked with `*' in the table). We argue that this is important because methods are in general prone to overfitting to this specific dataset. As an example, our result with pose prior shows increased error compared to our result without prior, although we find that pose priors helps to produce good surface structure and joint angles in the wild.

\textbf{Ablation Studies.} We investigate the importance of each dataset through ablation studies on Human3.6M. We compare the reconstruction error by training networks with: (1) Human3.6M; (2) Human3.6M and our captured dataset; and (3) Human3.6M, our captured dataset, and COCO. Note that setting (3) is the method we used for the previous comparison. We follow the same evaluation protocol and metric as in Table \ref{table:human3.6m}. The result is shown in Table \ref{table:ablation}. First, it is worth noting that with only Human3.6M as training data, we already achieve the best results among results marked with `*' in Table \ref{table:human3.6m}. Second, comparing (2) with (1), our new dataset provides an improvement despite the difference in background, human appearance and pose distribution between our dataset and Human3.6M. This verifies the value of our new dataset. Third, we see a drop in error when we add COCO to the training data, which suggests that our framework can take advantage of this dataset with only 2D human pose annotation for 3D pose estimation.


\begin{table}[t]
\centering
\begin{tabular}{p{6cm} c}
    \hline\addlinespace[1pt]
    Training data & MPJPE \\
    \hline\addlinespace[1pt]
    (1) Human3.6M & 65.6  \\
    (2) Human3.6M + Ours & 60.9 \\
    (3) Human3.6M + Ours + COCO & 58.3 \\
    \hline
\end{tabular}
\caption{Ablation studies on Human3.6M. The evaluation metric is Mean Per Joint Position Error in millimeter.}
\label{table:ablation}
\end{table}

\begin{figure}[t]
\centering
  \includegraphics[trim={0.1cm 0 0.7cm 0},clip,height=0.46\linewidth]{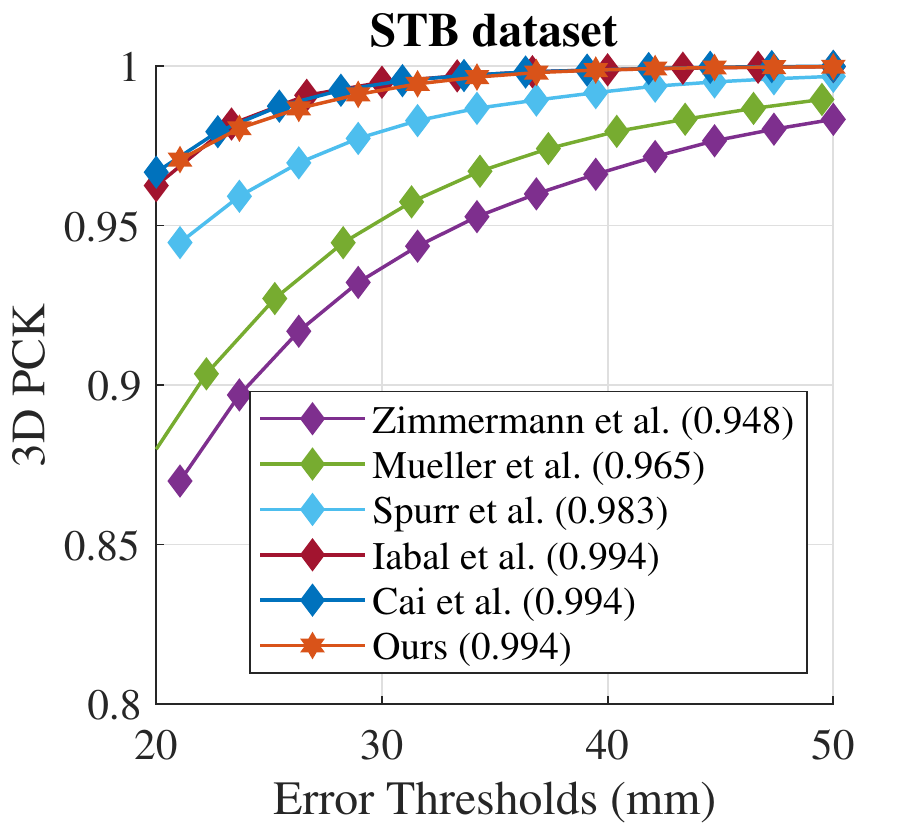}
  \includegraphics[trim={0.5cm 0 0.7cm 0},clip,height=0.46\linewidth]{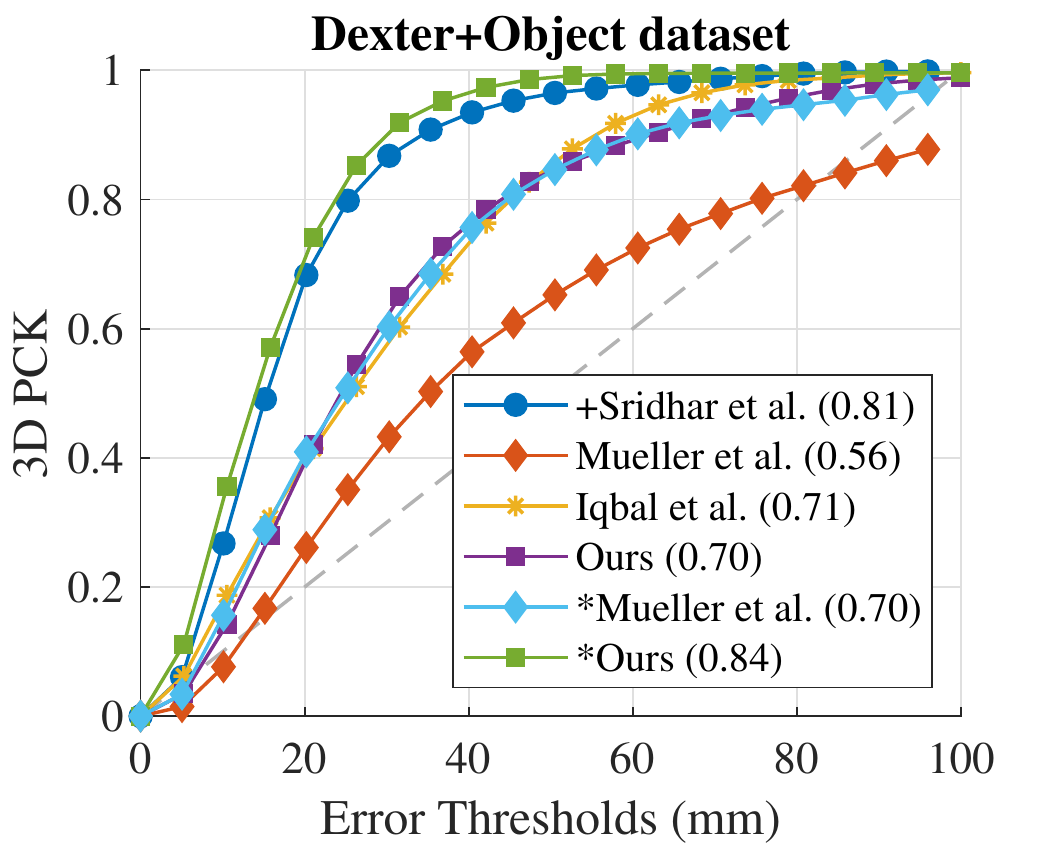}
\caption{Comparison with previous work on 3D hand pose estimation datasets. We plot PCK curve and show AUC in bracket for each method in legend. Left: results on the STB dataset \cite{zhang20163d} in 20mm - 50mm; right: results on Dexter+Object dataset \cite{sridhar2016real} in 0 - 100mm. Results with depth alignment are marked with `*'; the RGB-D based method is marked with `+'.}
\label{fig:STB}
\end{figure}

\subsubsection{3D Hand Pose Estimation.}
We evaluate our method on the Stereo Hand Pose Tracking Benchmark (STB) and Dexter+Object (D+O), and compare our result with previous methods. For this experiment we use the separate hand model of Frankenstein in \cite{joo2018}.

\textbf{STB.} Since the STB dataset has a palm joint rather than the wrist joint used in our method, we convert the palm joint to wrist joint as in \cite{zimmermann2017learning} to train our CNN. We also learn a linear regressor using the training set of STB dataset. During testing, we regress back the palm joint from our model fitting output for comparison. For the evaluation, we follow the previous work \cite{zimmermann2017learning} and compute the error after aligning the position of root joint and global scale with the ground truth, and report the Area Under Curve (AUC) of the Percentage of Correct Keypoints (PCK) curve in the 20mm-50mm range. The results are shown in the left of Fig.~\ref{fig:STB}. Our performance is on par with the state-of-the-art methods that are designed particularly for hand pose estimation. We also point out that the performance on this dataset has almost saturated, because the percentage is already above $90\%$ even at the lowest threshold.

\begin{figure}[t]
\centering
\begin{subfigure}{\linewidth}
  \centering
  \includegraphics[width=1.0\textwidth]{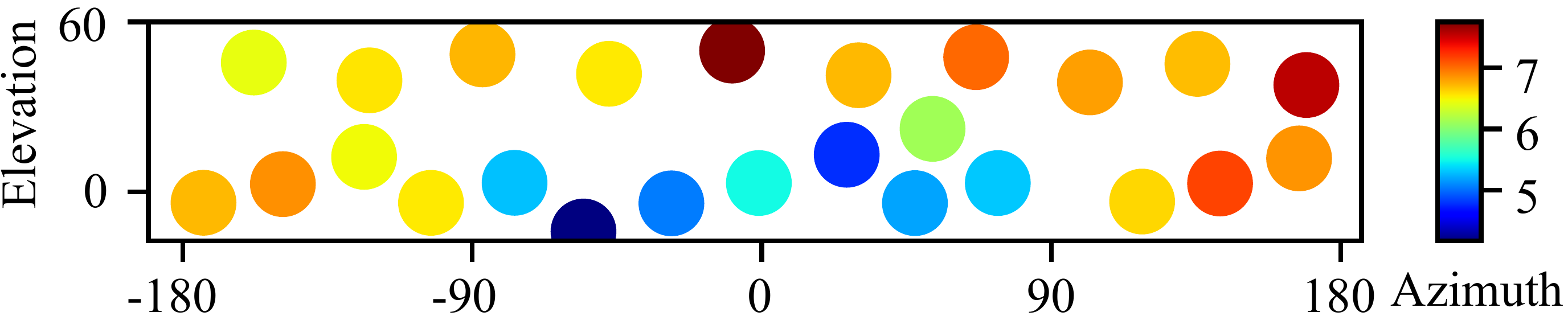}
\end{subfigure}
\begin{subfigure}{\linewidth}
  \centering
  \includegraphics[width=1.0\textwidth]{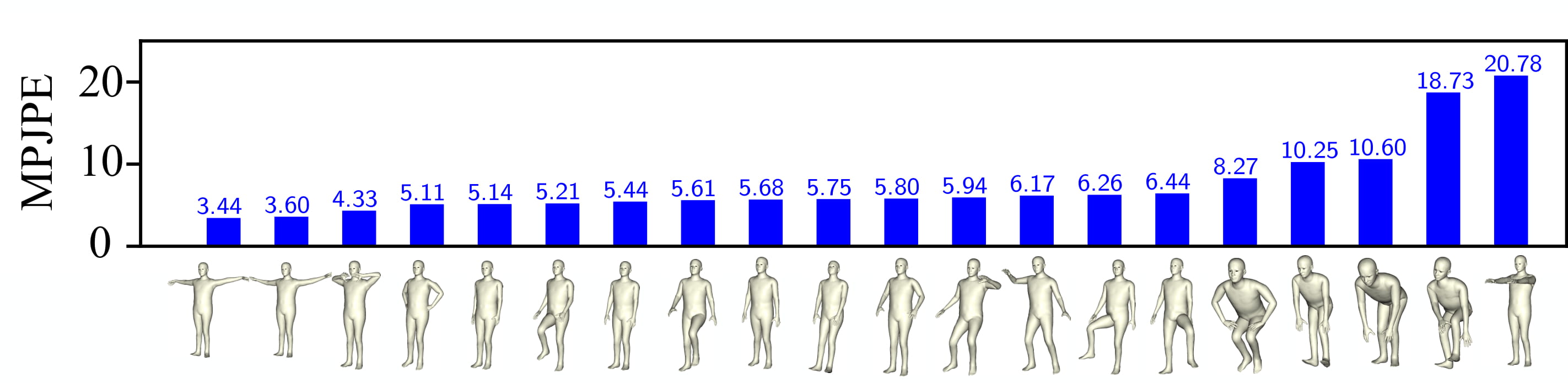}
\end{subfigure}
\caption{Evaluation result in Panoptic Studio. Top: accuracy vs. view point; bottom: accuracy vs. pose. The metric is MPJPE in cm. The average MPJPE for all testing samples is $6.30$ cm.}
\label{fig:panopticEvalution}
\end{figure}

\begin{figure}[t]
  \centering
  \includegraphics[width=1.0\linewidth]{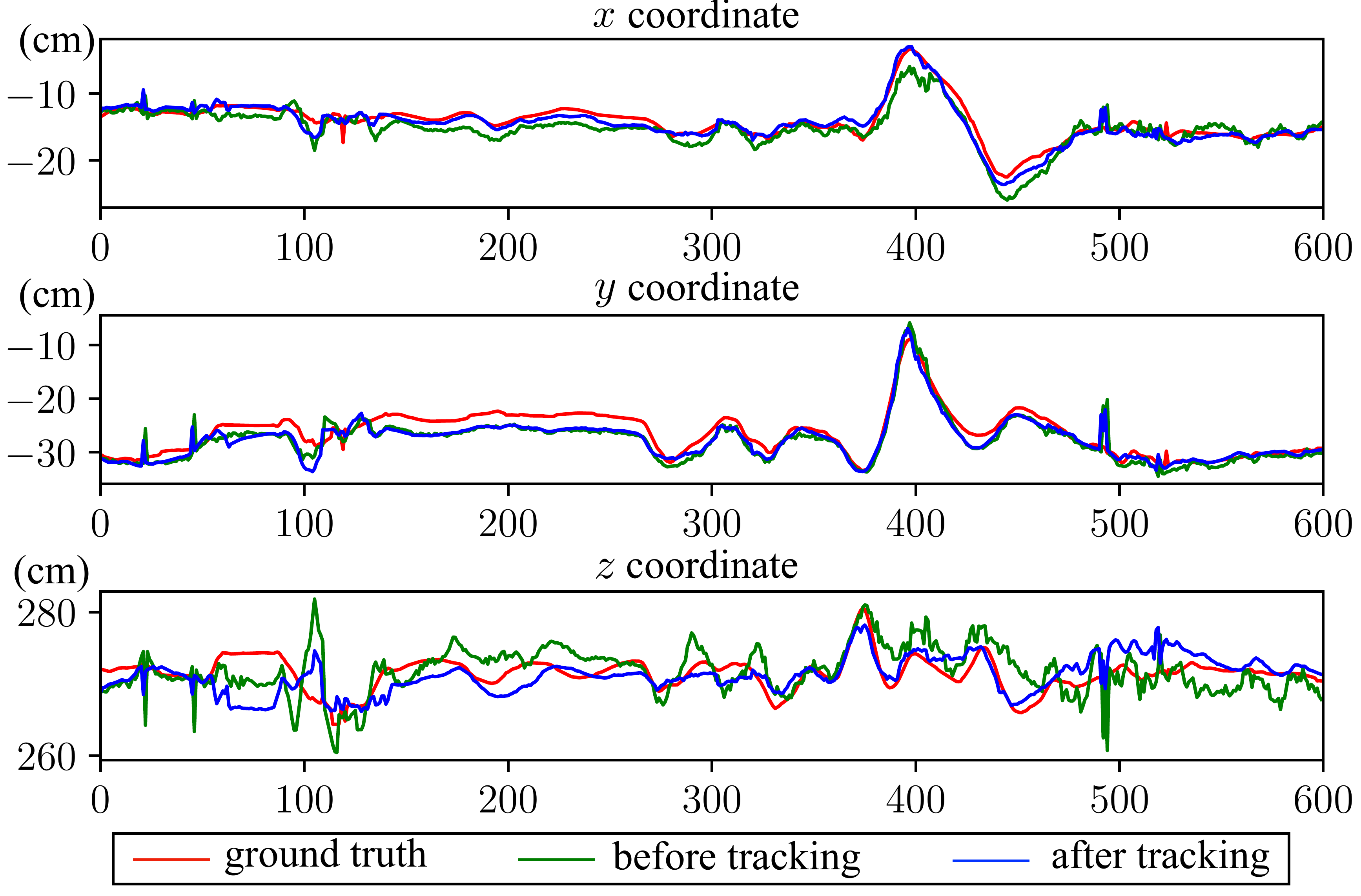}
\caption{The comparison of joint location across time before and after tracking with ground truth. The horizontal axes show frame numbers (30fps) and the vertical axes show joint locations in camera coordinate. The target joint here is the left shoulder of the subject.}
\label{fig:plot_tracking}
\end{figure}

\textbf{D+O.}  Following \cite{mueller2018ganerated} and \cite{Iqbal_2018_ECCV}, we report our results using a PCK curve and the corresponding AUC, as shown in the right of Fig.~\ref{fig:STB}. Since previous methods are evaluated by estimating the absolute 3D depth of 3D hand joints, we follow them by finding an approximate hand scale using a single frame in the dataset, and fix the scale during the evaluation. In this case, our performance (AUC=$0.70$) is comparable with the previous state-of-the-art \cite{Iqbal_2018_ECCV} (AUC=$0.71$). However, since there is fundamental depth-scale ambiguity for single-view pose estimation, we argue that aligning the root with the ground truth depth is a more reasonable evaluation setting. In this setting,  our method (AUC=$0.84$) outperforms the previous state-of-the-art method \cite{mueller2018ganerated} (AUC=$0.70$) in the same setting, and even achieves better performance than an RGB-D based method \cite{sridhar2016real} (AUC=$0.81$). 



\subsection{Quantitative Study for View and Pose Changes}
Our new 3D pose data contain multi-view images with the diverse body postures. This allows us to quantitatively study the performance of our method in view changes and body pose changes. We compare our single view 3D body reconstruction result with the ground truth. Due to the scale-depth ambiguity of monocular pose estimation, we align the depth of root joint to the ground truth by scaling our result along the ray directions from the camera center, and compute the Mean Per Joint Position Error (MPJPE) in centimeter. The average MPJPE for all testing samples is $6.30$ cm. We compute the average errors per each camera viewpoint, as shown in the top of Fig. \ref{fig:panopticEvalution}. Each camera viewpoint is represented by azimuth and elevation with respect to the subjects' initial body location. We reach two interesting findings: first, the performance worsens in the camera views with higher elevation due to the severe self-occlusion and foreshortening; second, the error is larger in back views compared to the frontal views because limbs are occluded by torso in many poses. At the bottom of Fig. \ref{fig:panopticEvalution}, we show the performance for varying body poses. We run k-means algorithm on the ground truth data to find body pose groups (the center poses are shown in the figure), and compute the error for each cluster. Body poses with more severe self-occlusion or foreshortening tend to have higher errors.

\subsection{The Effect of Mesh Tracking} To demonstrate the effect of our temporal refinement method, we compare the result of our method before and after this refinement stage using Panoptic Studio data. We plot the reconstructed left shoulder joint in Fig. \ref{fig:plot_tracking}.
We find that the result after tracking (in blue) tends to be more temporally stable than that before tracking (in green), and is often closer to the ground truth (in red). 

\subsection{Qualitative Evaluation}

\begin{figure*}[t]
  \centering
  \includegraphics[trim={0cm 0 0cm 12.7cm}, clip, width=\textwidth]{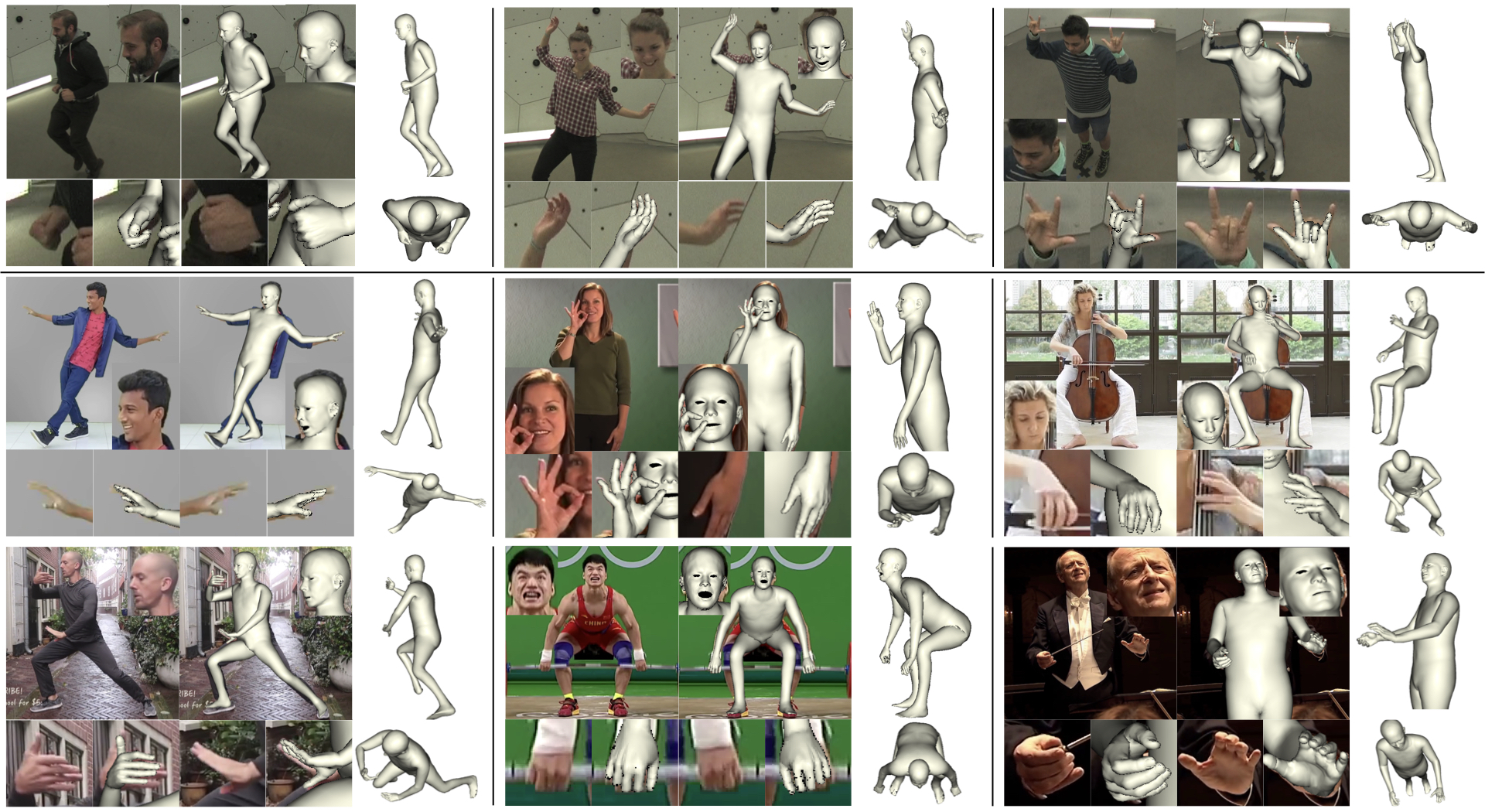}
  \caption{Qualitative results of our method on in-the-wild images. For each example, we show input images and our prediction with zoom-in views as well as side and top views.}
  \label{fig:Q2}
\end{figure*}

\noindent \textbf{Qualitative Results on Images:} In this section we present qualitative results of our method on individual images in Fig. \ref{fig:Q2}. We show results on images with various background, human appearance and poses. Our method works well for both indoor Mocap images (the first row in Fig. \ref{fig:Q2}) and in-the-wild images (the latter 2 rows).

\noindent \textbf{Qualitative Results on Video Sequences:} We show results of our method on video sequences. We test our method on two kinds of videos. First, we take videos of human motion using camera by ourselves; second, we use videos downloaded from Youtube. The results are presented in our supplementary video. For videos where only the upper body of the target person is visible, we assume that the orientation of torso and legs is vertically downward in Equation \ref{eq:modelfitting_BOF}.

\section{Discussion}
In this paper, we present a method to simultaneously reconstruct 3D total motion of a single person from an image or a monocular video. We thoroughly evaluate the robustness of our method on various benchmarks and demonstrate monocular 3D total motion capture results on in-the-wild videos. There are some limitations with our method. First, we observe failure cases when a significant part of the target person is invisible (out of image boundary or occluded by other objects) due to erroneous network prediction. Second, our hand pose detector fails in the case of insufficient resolution or severe motion blur. Third, our CNN requires bounding boxes for body and hands as input, and cannot handle multiple bodies or hands simultaneously. Solving these problems points to interesting future directions.



%% file: supplementary.tex
\noindent{\Large\textbf{Appendix.}}

\section{New 3D Human Pose Dataset}

In this section, we provide more details of the new 3D human pose dataset that we collect.

\subsection{Methodology}

We build this dataset in 3 steps:
\begin{itemize}
    \item We randomly recruit 40 volunteers on campus and capture their motion in a multi-view system \cite{joo2015panoptic, joo2017panoptic}. During the capture, all subjects follow the motion in the same video of around 2.5 minutes recorded in advance.
    \item We use multi-view 3D reconstruction algorithms \cite{joo2015panoptic, joo2017panoptic, simon2017hand} to reconstruct 3D body, hand and face keypoints.
    \item We run filters on the reconstruction results. We compute the average lengths of all bones for every subject, and discard a frame if the difference between the length of any bone in the frame and the average length is above a certain threshold. We further manually verify the correctness of hand annotations by projecting the skeletons onto 3 camera views and checking the alignment between the projection and images.
\end{itemize}

\subsection{Statistics and Examples}

\begin{figure}[t]
  \centering
  \includegraphics[trim={22cm 0 18cm 8cm},clip,width=0.495\linewidth]{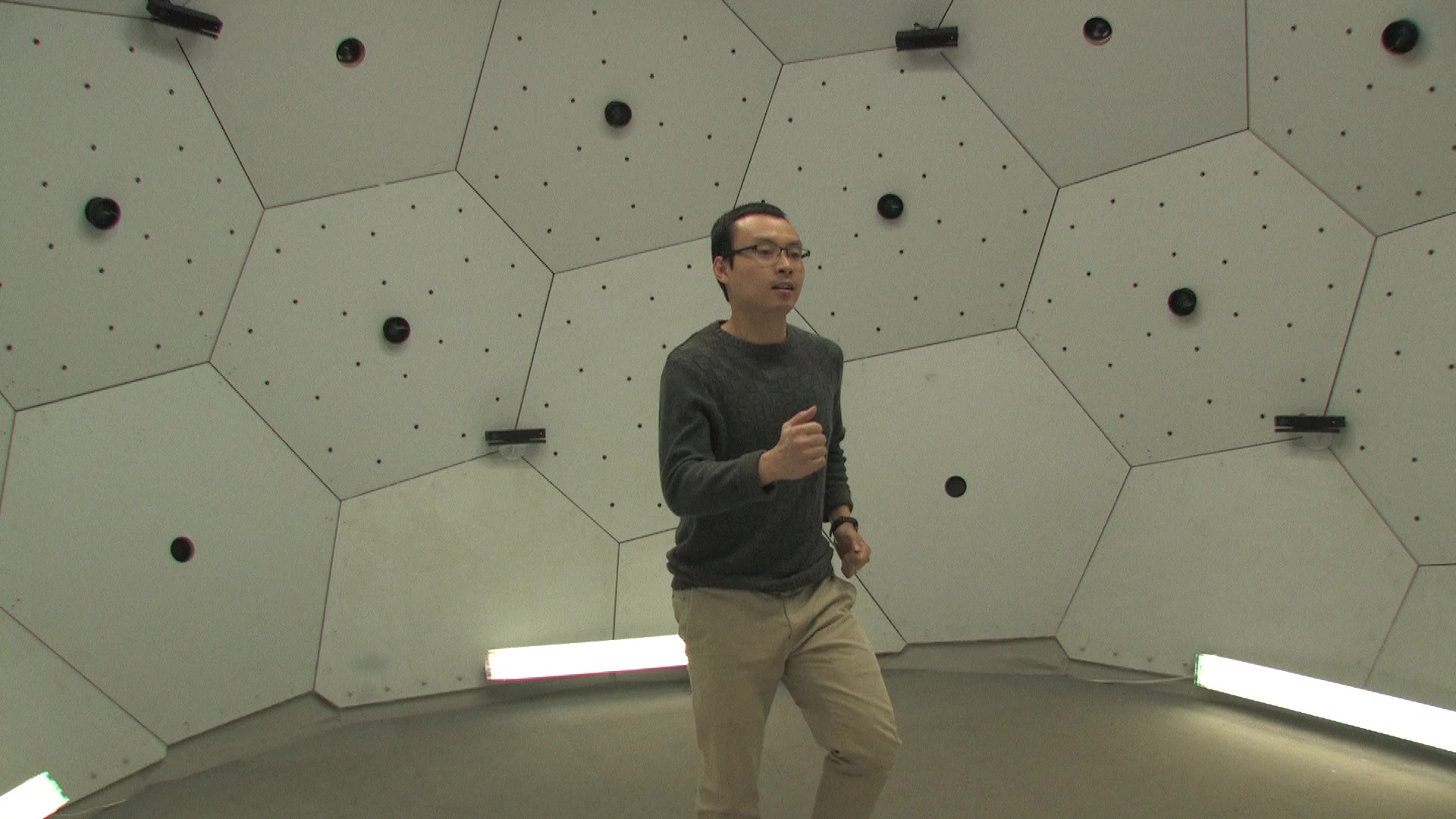}
  \includegraphics[trim={22cm 0 18cm 8cm},clip,width=0.495\linewidth]{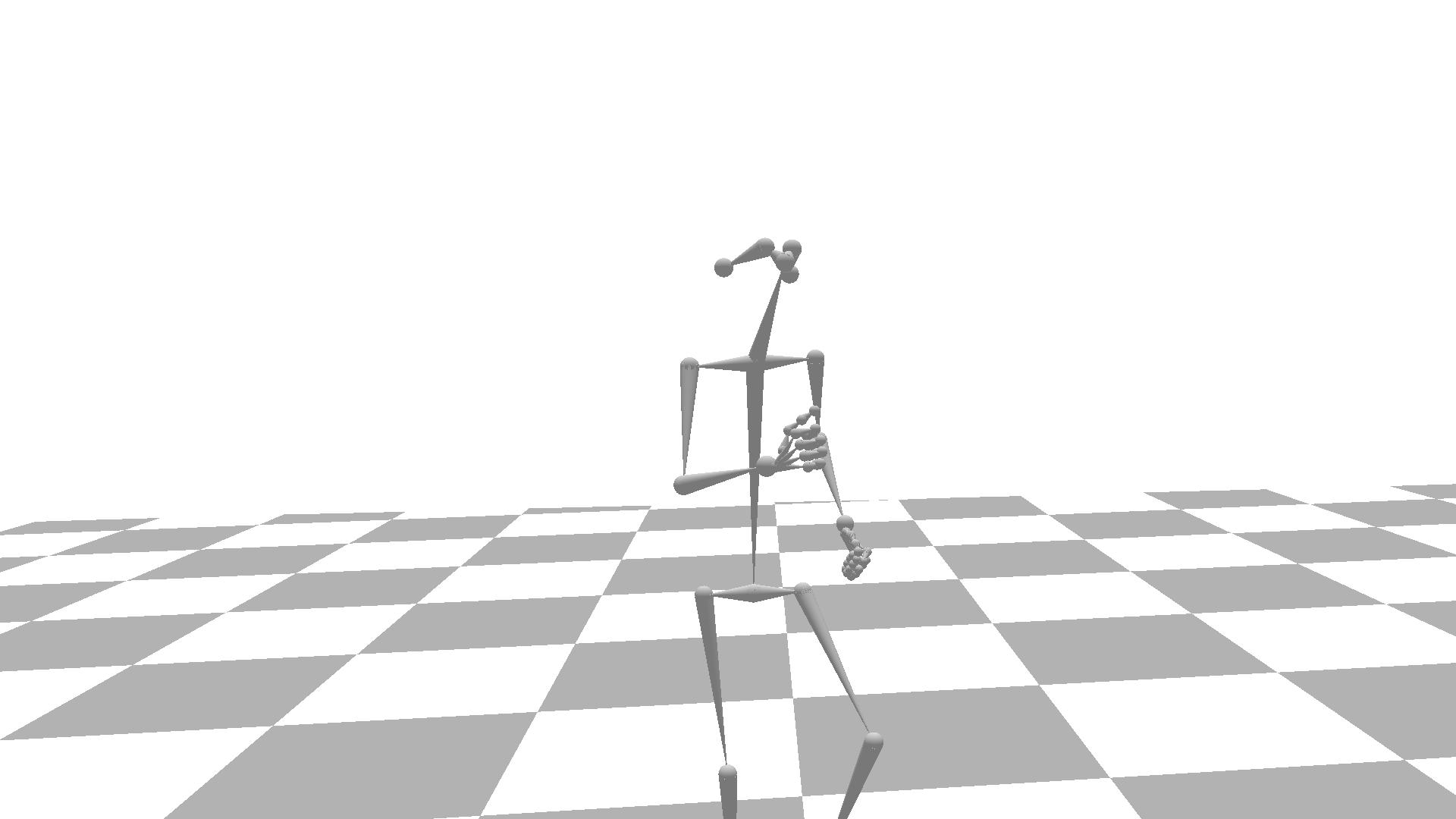}
  \includegraphics[trim={22cm 0 18cm 8cm},clip,width=0.495\linewidth]{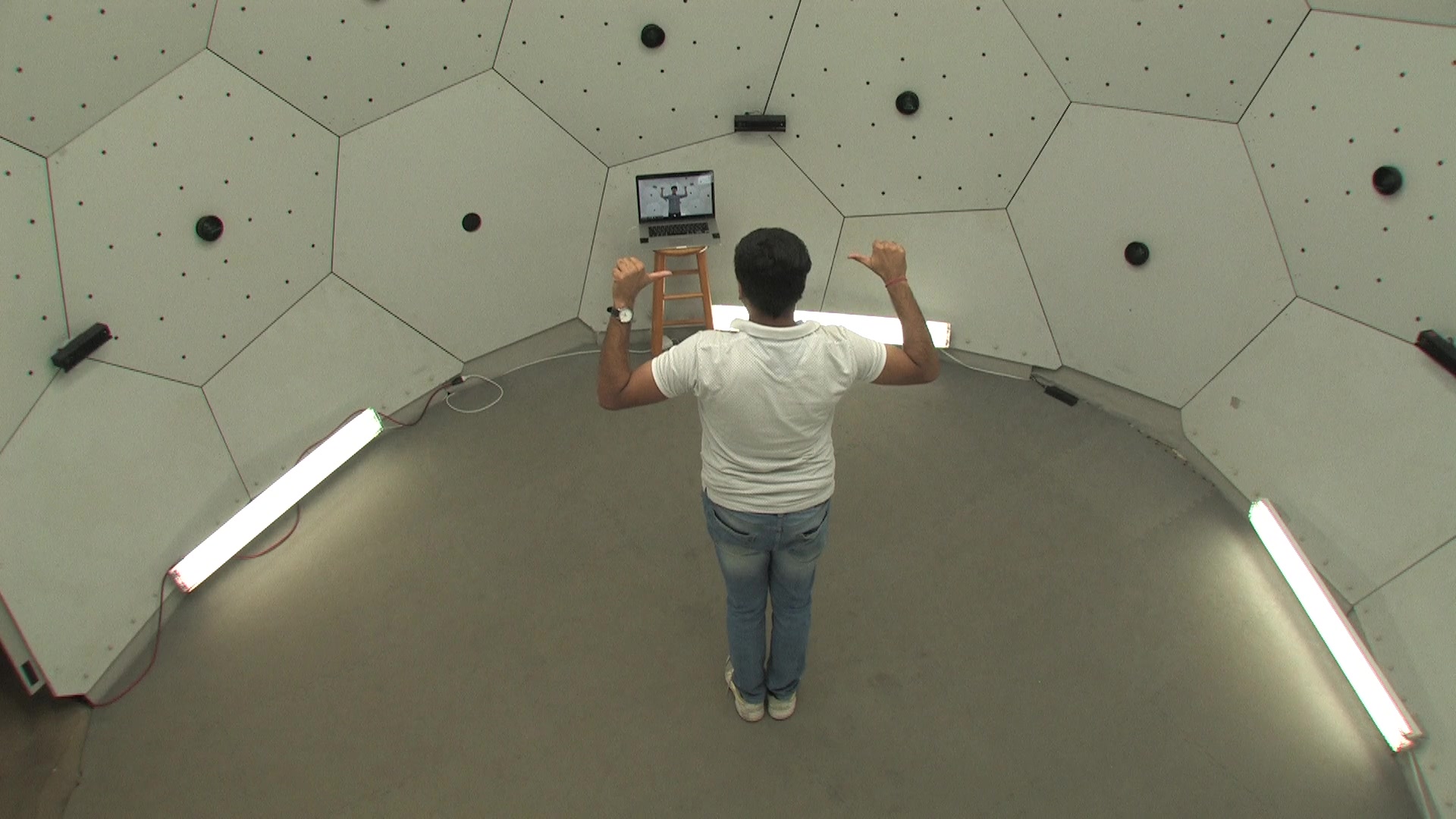}
  \includegraphics[trim={22cm 0 18cm 8cm},clip,width=0.495\linewidth]{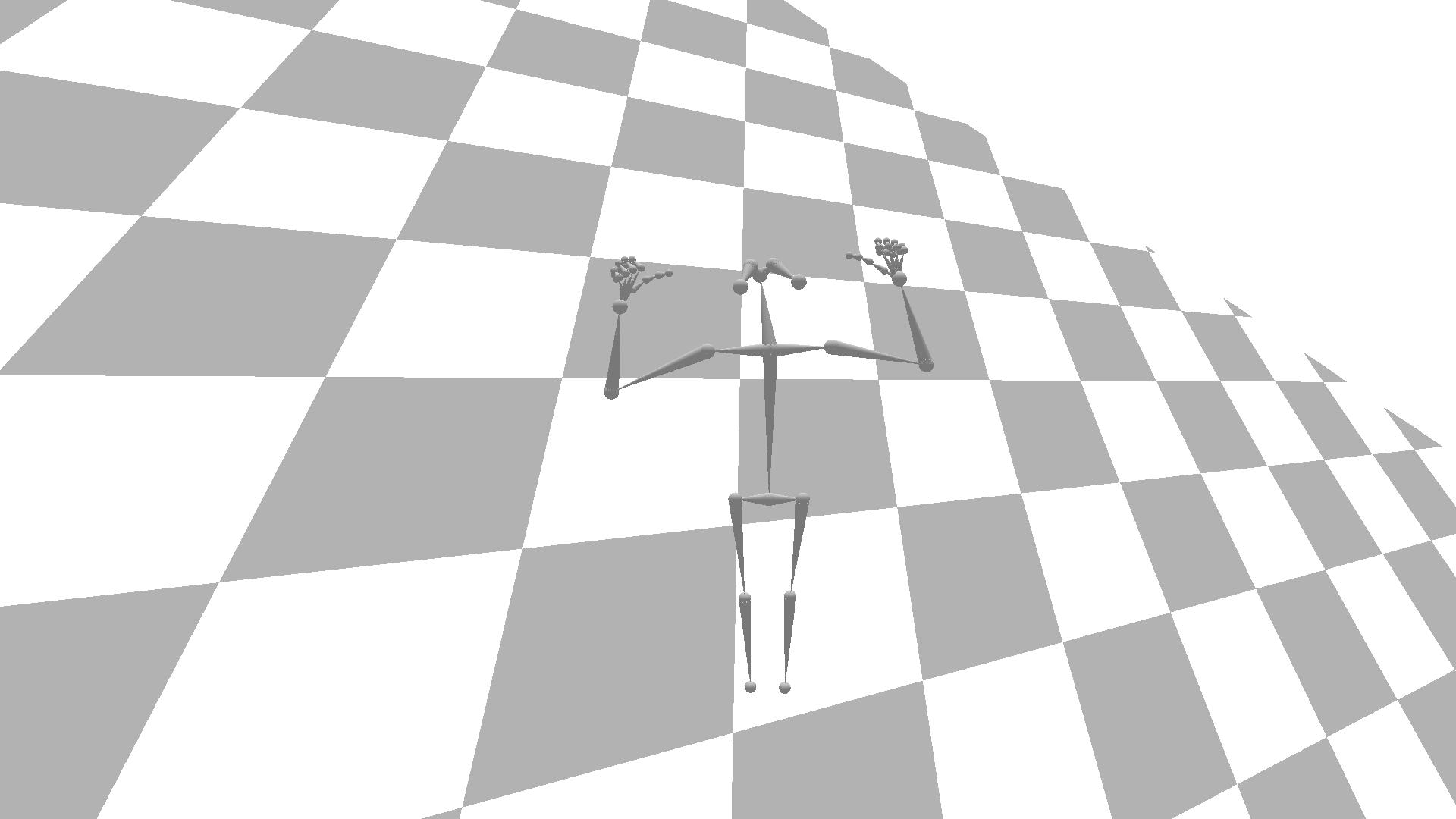}
\caption{Example images and 3D annotations from our new 3D human pose dataset.}
\label{fig:dataset}
\end{figure}

To train our networks, we use our captured 3D body data and hand data, include a total of \textbf{834K} image-annotation pairs for bodies and \textbf{111K} pairs for hands. Example data are shown in Fig. \ref{fig:dataset} and our supplementary video.

\section{Network Skeleton Definition}

\begin{figure*}[t]
  \centering
  \includegraphics[width=1.0\linewidth]{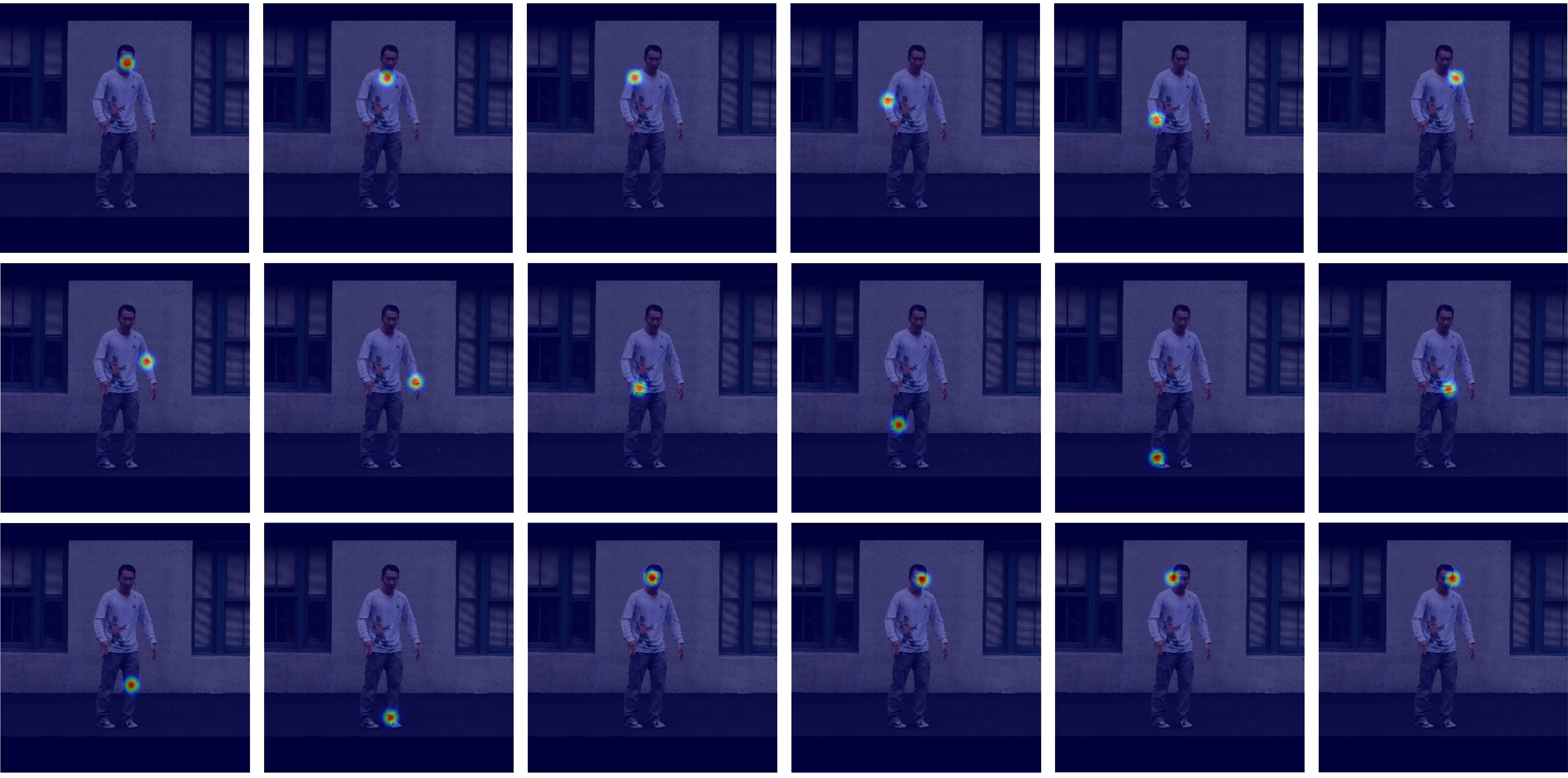}
\caption{Joint confidence maps predicted by our CNN for a body image.}
\label{fig:kps}
\end{figure*}
\begin{figure*}[t]
  \centering
  \includegraphics[width=0.79\linewidth]{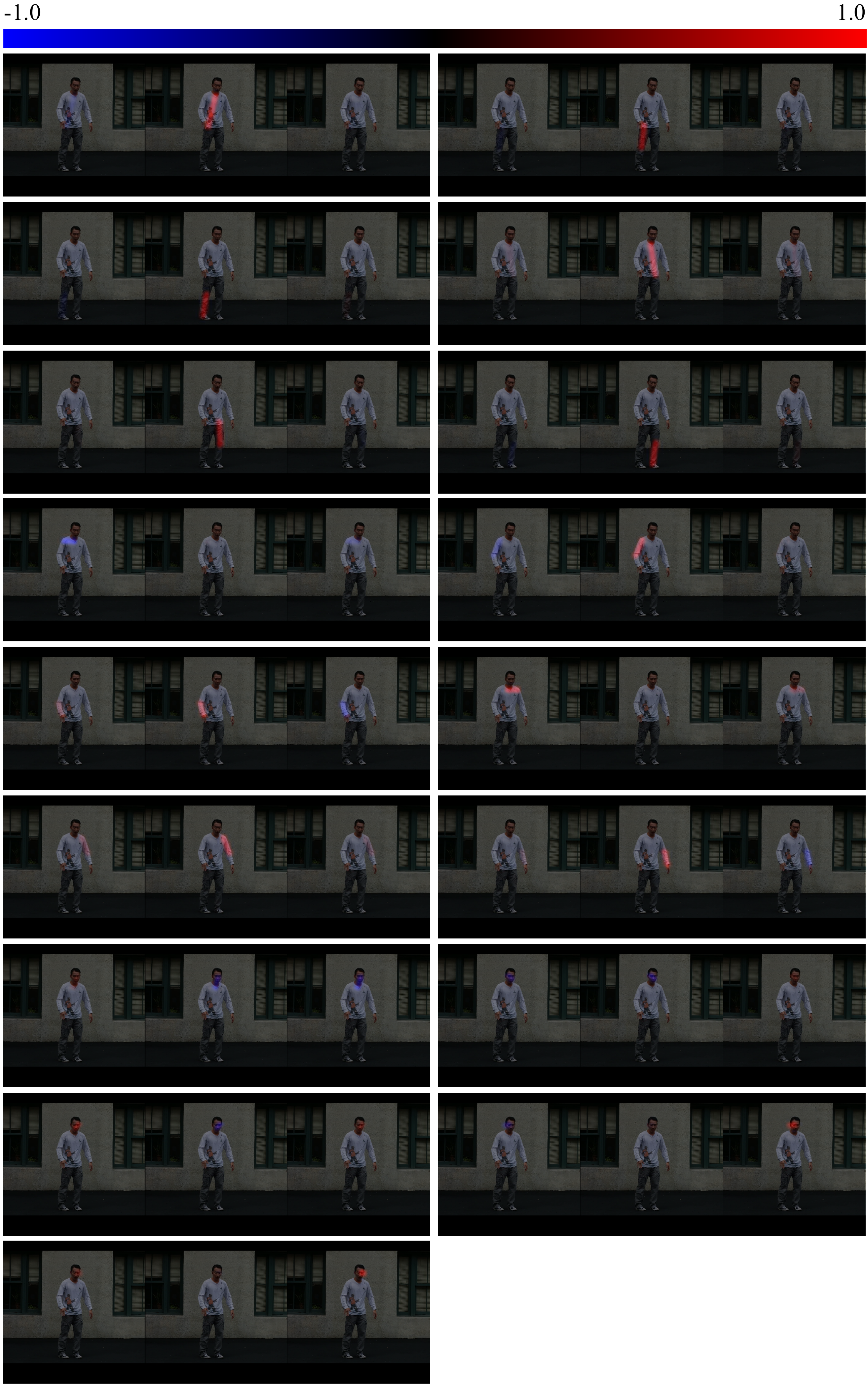}
\caption{Part Orientation Fields predicted by our CNN for a body image. For each body part we visualize $x, y, z$ channels separately.}
\label{fig:pof}
\end{figure*}
\begin{figure*}[t]
  \centering
  \includegraphics[width=1.0\linewidth]{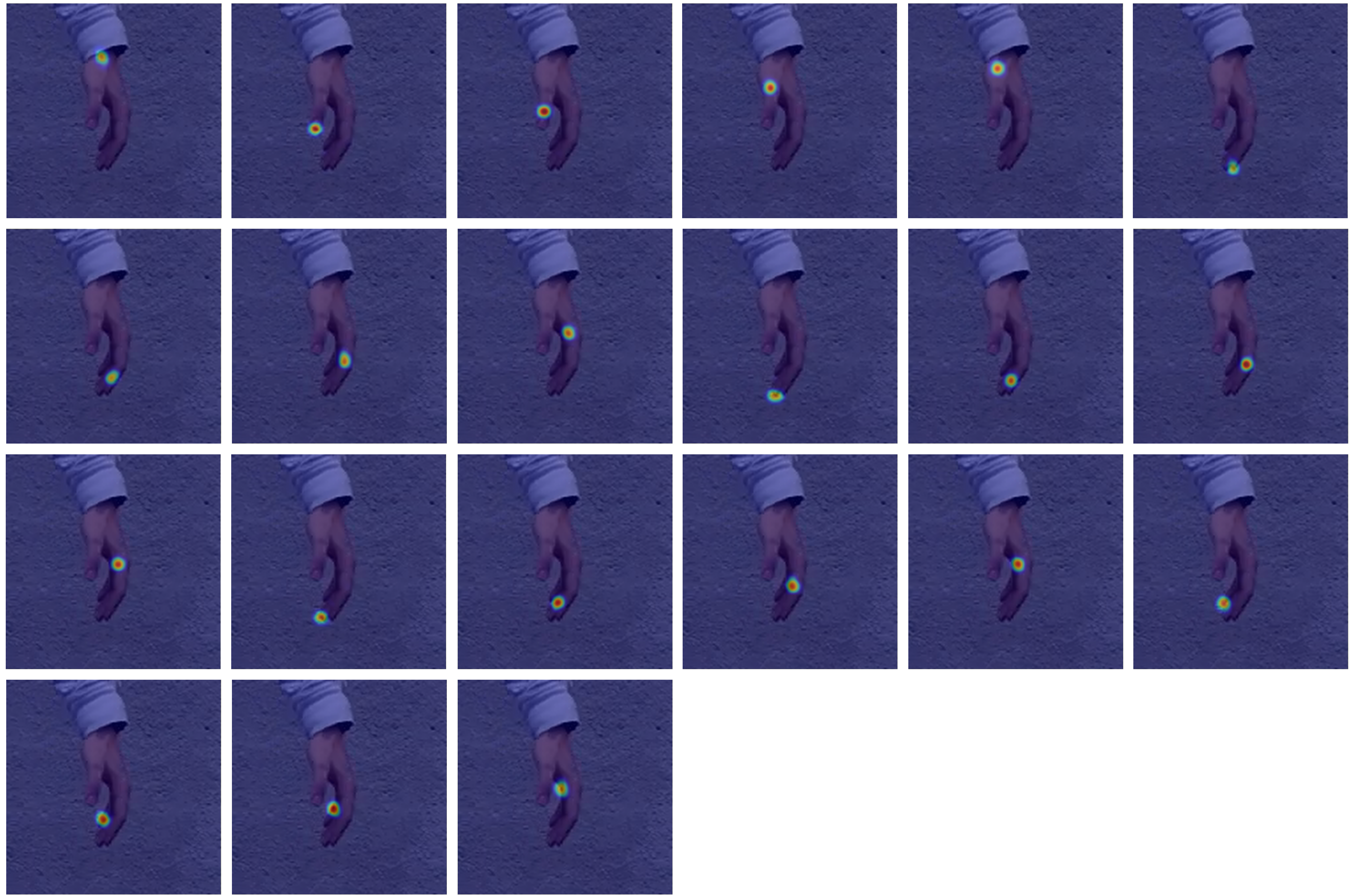}
\caption{Joint confidence maps predicted by our CNN for a hand image.}
\label{fig:handkps}
\end{figure*}
\begin{figure*}[t]
  \centering
  \includegraphics[width=0.71\linewidth]{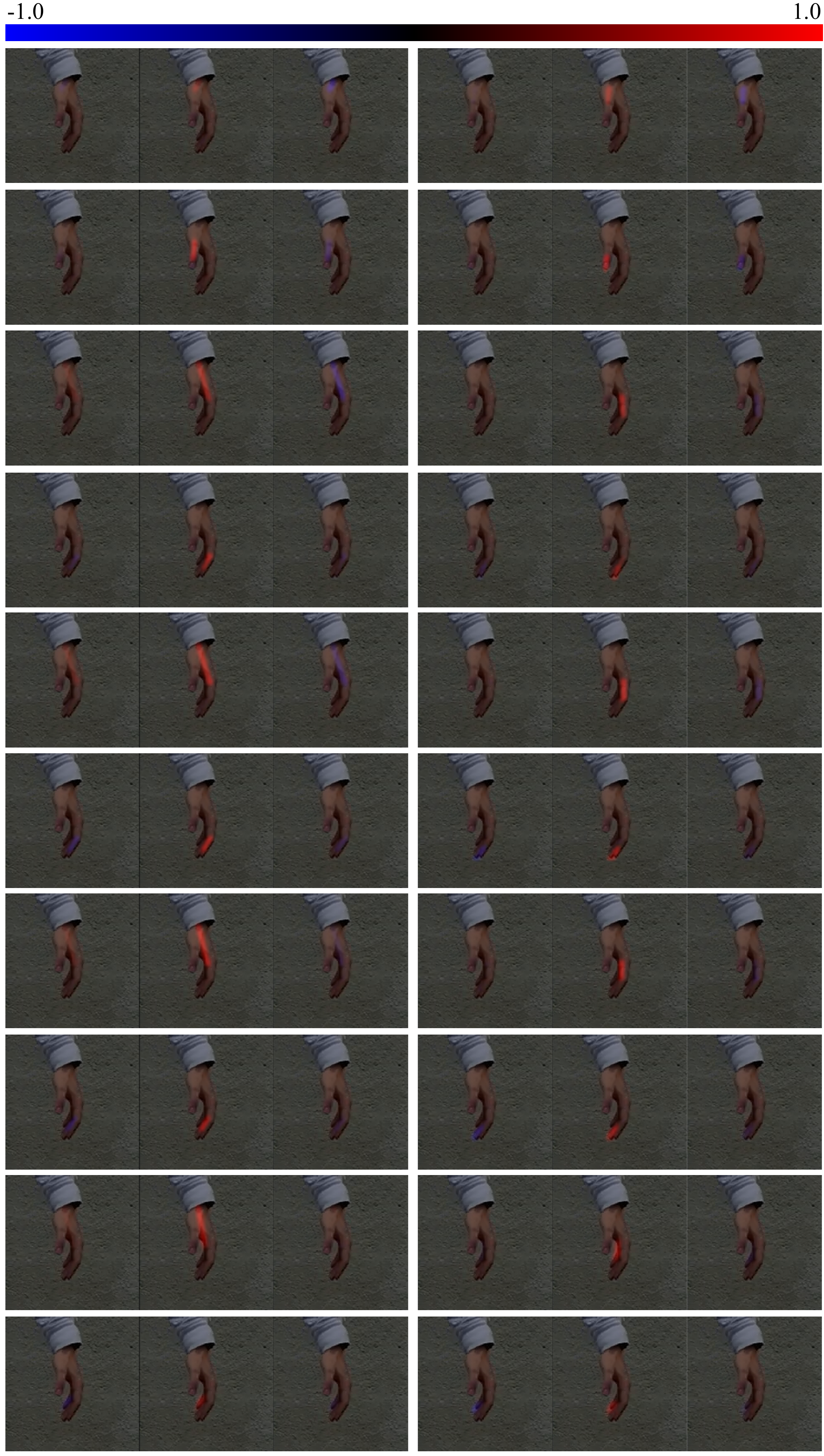}
\caption{Part Orientation Fields predicted by our CNN for a hand image. For each hand part we visualize $x, y, z$ channels separately.}
\label{fig:handpof}
\end{figure*}

In this section we specify the skeleton hierarchy $\mathbb S$ we use for our Part Orientation Fields and joint confidence maps. As shown in Fig. \ref{fig:skeleton}, we predict 18 keypoints for the body and POFs for 17 body parts, so $\mathbf S^B \in \mathbb R^{18 \times 368 \times 368}, \mathbf L^B \in \mathbb R^{51 \times 368 \times 368}$. Analogously, we predict 21 joints for each hand and POFs for 20 hand parts, so $\mathbf S^{LH}$ and $\mathbf S^{RH}$ have the dimension ${21 \times 368 \times 368}$, while $\mathbf L^{LH}$ and $\mathbf L^{RH}$ have the dimension $60 \times 368 \times 368$. Note that we train a CNN only for left hands, and we horizontally flip images of right hands before they are fed into the network during testing. Some example outputs of our CNN are shown in Fig. \ref{fig:kps}, \ref{fig:pof}, \ref{fig:handkps}, \ref{fig:handpof}.

\section{Deformable Human Model}

\subsection{Model Parameters}
As explained in the main paper, we use Adam model introduced in \cite{joo2018} for total body motion capture. The model parameters $\boldsymbol\Psi$ include the shape parameters $\boldsymbol{\phi} \in \mathbb{R}^{K_\phi}$, where $K_\phi=30$ is the dimension of shape deformation space, the pose parameters $\boldsymbol{\theta} \in \mathbb{R}^{J \times 3}$ where the $J=62$ is the number of joints in the model\footnote{The model has 22 body joints and 20 joints for each hand.}, the global translation parameters $\boldsymbol{t} \in \mathbb R^3$, and the facial expression parameter $\boldsymbol{\sigma} \in \mathbb{R}^{K_\sigma}$ where $K_\sigma=200$ is the number of facial expression bases.

\subsection{3D Keypoints Definition}

In this section we specify the correspondences between the keypoints predicted by our networks and Adam keypoints.

\begin{figure}[t]
  \centering
  \includegraphics[width=1.0\linewidth]{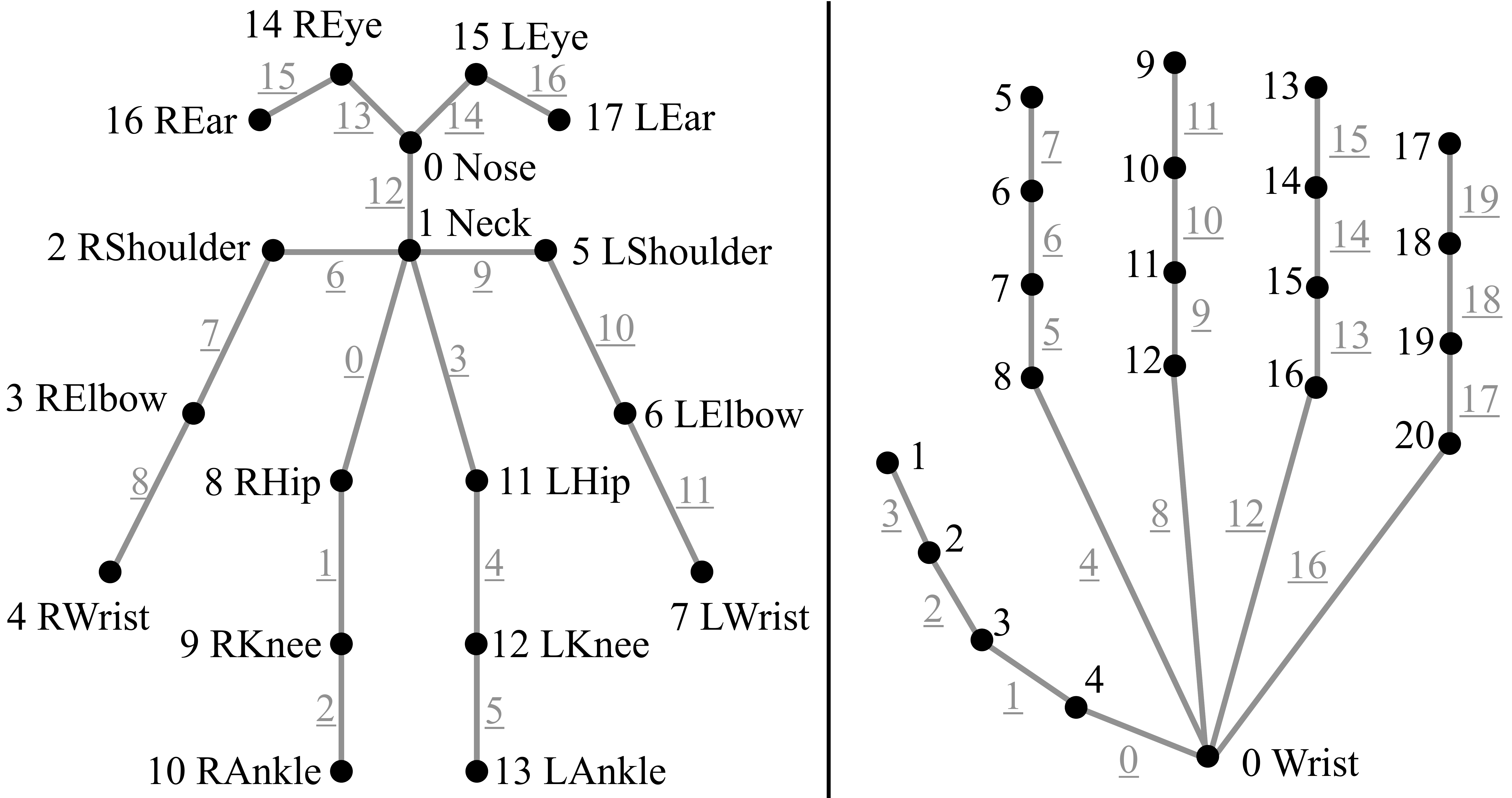}
\caption{Illustration on the skeleton hierarchy $\mathbb S$ in our POFs and joint confidence maps. The joints are shown in black, and body parts for POFs are shown in gray with indices underlined. On the left we show the skeleton used in our body network; on the right we show the skeleton used in our hand network.}
\label{fig:skeleton}
\end{figure}

Regressors for the body are directly provided by \cite{joo2018}, which define keypoints as linear combination of mesh vertices. During mesh fitting (Section 5 of the main paper), given current mesh $M(\boldsymbol{\Psi})$ determined by mesh parameters $\boldsymbol{\Psi} = (\boldsymbol{\phi}, \boldsymbol{\theta}, \boldsymbol{t}, \boldsymbol{\sigma})$, we use these regressors to compute joints $\{\tilde{\mathbf J}^B_m\}$ from the mesh vertices, and further $\{\tilde{\mathbf P}^B_{(m,n)}\}$ by Equation 1 in the main paper. $\{\tilde{\mathbf J}^B_m\}$ and $\{\tilde{\mathbf P}^B_{(m,n)}\}$ follow the skeleton structure in Fig. \ref{fig:skeleton}. $\{\tilde{\mathbf J}^B_m\}$ and $\{\tilde{\mathbf P}^B_{(m,n)}\}$ are used in Equation 4 and 5 in the main paper respectively to fit the body pose.

Joo \etal \cite{joo2018} also provides regressors for both hands, so we follow the same setup as body to define keypoints and hand parts $\{\tilde{\mathbf J}^{LH}_m\}, \{\tilde{\mathbf J}^{RH}_m\}, \{\tilde{\mathbf P}^{LH}_{(m,n)}\}, \{\tilde{\mathbf P}^{RH}_{(m,n)}\}$, which are used in Equation 7 in the main paper to fit hand pose. Note that the wrists appear in both skeletons of Fig. \ref{fig:skeleton}, so actually $\tilde{\mathbf J}^{LH}_0 = \tilde{\mathbf J}^{B}_7, \tilde{\mathbf J}^{RH}_0 = \tilde{\mathbf J}^{B}_4$. We only use 2D keypoint constraints from the body network, i.e., $\mathbf j^B_4, \mathbf j^B_7$ in Equation 4, ignoring the keypoint measurements from hand network $\mathbf j^{LH}_0$ and $\mathbf j^{RH}_0$ in Equation 7, since the body network is usually more stable in output.

\begin{figure}[t]
  \centering
  \includegraphics[width=\linewidth]{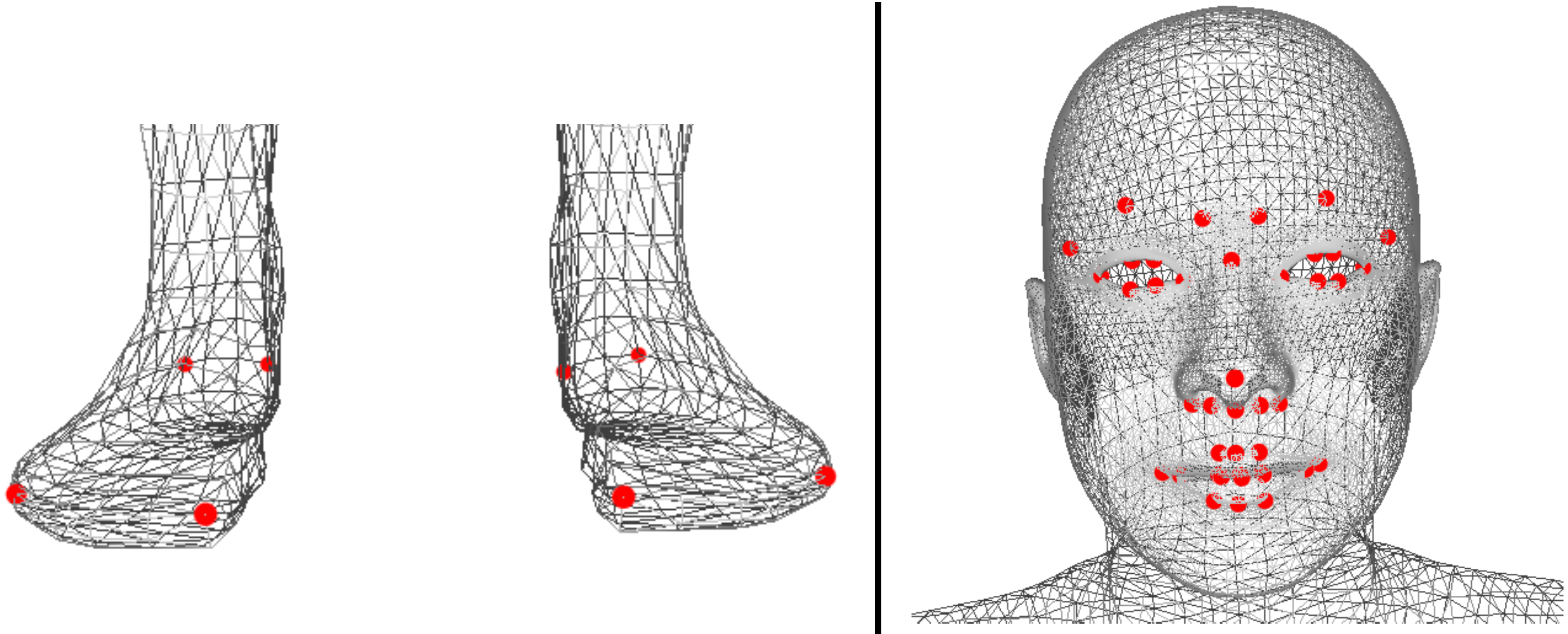}
\caption{We plot Adam vertices used as keypoints for mesh fitting in red dots. Left: vertices used to fit both feet (the middle points between the 2 vertices at the back are  keypoints); right: vertices used to fit facial expression.}
\label{fig:footface}
\end{figure}

For Equation 8 in the main paper, we use 2D foot keypoint locations from OpenPose as $\{\mathbf j^T_m\}$, including big toes, small toes and heels of both feet. On the Adam side, we directly use mesh vertices as keypoints $\{\tilde{\mathbf J}^T_m\}$ for big toes and small toes on both feet. We use the middle point between a pair of vertices at the back of each feet as the heel keypoint,
as shown in Fig. \ref{fig:footface} (left).

In order to get facial expression, we also directly fit Adam vertices using the 2D face keypoints predicted by OpenPose (Equation 9 in the main paper). Note that although OpenPose provides 70 face keypoints, we only use 41 keypoints on eyes, nose, mouth and eyebrows, ignoring those on the face contour. The Adam vertices used for fitting are illustrated in Fig. \ref{fig:footface} (right).

\section{Implmentation Details}

In this section, we provide details about the parameters we use in our implementation.

In Equation 4 and 5 of the main paper, we use $$w^B_{\text{POF}} = 22500, w_p^B = 200.$$
We have similarly defined weights for left and right hands omitted in Equation 7, for which we use $$w^{LH}_{\text{POF}} = w^{RH}_{\text{POF}} = 2500, w_p^{LH} = w_p^{RH} = 10.$$
Weights for Equation 10 (omitted in the main paper) are $$ w^{\boldsymbol{\phi}} = 0.01, w^{\boldsymbol{\sigma}} = 100. $$
In Equation 15, a balancing weight is omitted for which we use $$w_{\Delta z} = 0.25.$$
In Equation 16, $\mathcal F_{\text{POF}}$ consists of POF terms for body, left hands and right hands, i.e., $\mathcal F_{\text{POF}} = \mathcal F^B_{\text{POF}} + \mathcal F^{LH}_{\text{POF}} + \mathcal F^{RH}_{\text{POF}}$. We use weights $25, 1, 1$ to balance these 3 terms.

%% file: 3DPAF-arxiv.bbl
\begin{thebibliography}{10}\itemsep=-1pt

\bibitem{CMUMocap}
Cmu motion capture database.
\newblock \url{http://mocap.cs.cmu.edu/resources.php}.

\bibitem{VICON}
Vicon motion systems.
\newblock \url{www.vicon.com}.

\bibitem{ceres-solver}
S.~Agarwal, K.~Mierle, and Others.
\newblock Ceres solver.
\newblock \url{http://ceres-solver.org}.

\bibitem{akhter2015pose}
I.~Akhter and M.~J. Black.
\newblock Pose-conditioned joint angle limits for 3d human pose reconstruction.
\newblock In {\em CVPR}, 2015.

\bibitem{Andriluka-14}
M.~Andriluka, L.~Pishchulin, P.~Gehler, and B.~Schiele.
\newblock {2D} human pose estimation: New benchmark and state of the art
  analysis.
\newblock In {\em CVPR}, 2014.

\bibitem{anguelov2005scape}
D.~Anguelov, P.~Srinivasan, D.~Koller, S.~Thrun, J.~Rodgers, and J.~Davis.
\newblock Scape: shape completion and animation of people.
\newblock {\em TOG}, 2005.

\bibitem{Baak2011}
A.~Baak, M.~M, G.~Bharaj, H.-p. Seidel, and C.~Theobalt.
\newblock {A Data-Driven Approach for Real-Time Full Body Pose Reconstruction
  from a Depth Camera}.
\newblock In {\em ICCV}, 2011.

\bibitem{blanz1999morphable}
V.~Blanz and T.~Vetter.
\newblock A morphable model for the synthesis of 3d faces.
\newblock In {\em Proceedings of the 26th annual conference on Computer
  graphics and interactive techniques}, pages 187--194. ACM
  Press/Addison-Wesley Publishing Co., 1999.

\bibitem{Bogo2016}
F.~Bogo, A.~Kanazawa, C.~Lassner, P.~Gehler, J.~Romero, and M.~J. Black.
\newblock {Keep it SMPL: Automatic Estimation of 3D Human Pose and Shape from a
  Single Image}.
\newblock In {\em ECCV}, 2016.

\bibitem{bulat2016human}
A.~Bulat and G.~Tzimiropoulos.
\newblock Human pose estimation via convolutional part heatmap regression.
\newblock In {\em ECCV}, 2016.

\bibitem{Cai_2018_ECCV}
Y.~Cai, L.~Ge, J.~Cai, and J.~Yuan.
\newblock Weakly-supervised 3d hand pose estimation from monocular rgb images.
\newblock In {\em ECCV}, 2018.

\bibitem{cao2014facewarehouse}
C.~Cao, Y.~Weng, S.~Zhou, Y.~Tong, and K.~Zhou.
\newblock Facewarehouse: A 3d facial expression database for visual computing.
\newblock {\em TVCG}, 2014.

\bibitem{cao2017realtime}
Z.~Cao, T.~Simon, S.-E. Wei, and Y.~Sheikh.
\newblock Realtime multi-person 2d pose estimation using part affinity fields.
\newblock In {\em CVPR}, 2017.

\bibitem{Chen2017}
C.-H. Chen and D.~Ramanan.
\newblock {3D Human Pose Estimation = 2D Pose Estimation + Matching}.
\newblock In {\em CVPR}, 2017.

\bibitem{dabral2018learning}
R.~Dabral, A.~Mundhada, U.~Kusupati, S.~Afaque, A.~Sharma, and A.~Jain.
\newblock Learning 3d human pose from structure and motion.
\newblock In {\em ECCV}, 2018.

\bibitem{Elhayek-15}
A.~Elhayek, E.~Aguiar, A.~Jain, J.~Tompson, L.~Pishchulin, M.~Andriluka,
  C.~Bregler, B.~Schiele, and C.~Theobalt.
\newblock Efficient convnet-based marker-less motion capture in general scenes
  with a low number of cameras.
\newblock In {\em CVPR}, 2015.

\bibitem{AAAI1816471}
H.~Fang, Y.~Xu, W.~Wang, X.~Liu, and S.-C. Zhu.
\newblock Learning pose grammar to encode human body configuration for 3d pose
  estimation.
\newblock In {\em AAAI}, 2018.

\bibitem{Gall-09}
J.~Gall, C.~Stoll, E.~De~Aguiar, C.~Theobalt, B.~Rosenhahn, and H.-P. Seidel.
\newblock Motion capture using joint skeleton tracking and surface estimation.
\newblock In {\em CVPR}, 2009.

\bibitem{h36m_pami}
C.~Ionescu, D.~Papava, V.~Olaru, and C.~Sminchisescu.
\newblock Human3.6m: Large scale datasets and predictive methods for 3d human
  sensing in natural environments.
\newblock {\em TPAMI}, 2014.

\bibitem{Iqbal_2018_ECCV}
U.~Iqbal, P.~Molchanov, T.~Breuel Juergen~Gall, and J.~Kautz.
\newblock Hand pose estimation via latent 2.5d heatmap regression.
\newblock In {\em ECCV}, 2018.

\bibitem{joo2015panoptic}
H.~Joo, H.~Liu, L.~Tan, L.~Gui, B.~Nabbe, I.~Matthews, T.~Kanade, S.~Nobuhara,
  and Y.~Sheikh.
\newblock Panoptic studio: A massively multiview system for social motion
  capture.
\newblock In {\em CVPR}, 2015.

\bibitem{joo2017panoptic}
H.~Joo, T.~Simon, X.~Li, H.~Liu, L.~Tan, L.~Gui, S.~Banerjee, T.~Godisart,
  B.~Nabbe, I.~Matthews, et~al.
\newblock Panoptic studio: A massively multiview system for social interaction
  capture.
\newblock {\em TPAMI}, 2017.

\bibitem{joo2018}
H.~Joo, T.~Simon, and Y.~Sheikh.
\newblock Total capture: A 3d deformation model for tracking faces, hands, and
  bodies.
\newblock In {\em CVPR}, 2018.

\bibitem{kanazawa2018end}
A.~Kanazawa, M.~J. Black, D.~W. Jacobs, and J.~Malik.
\newblock End-to-end recovery of human shape and pose.
\newblock In {\em CVPR}, 2018.

\bibitem{lin2014microsoft}
T.-Y. Lin, M.~Maire, S.~Belongie, J.~Hays, P.~Perona, D.~Ramanan,
  P.~Doll{\'a}r, and C.~L. Zitnick.
\newblock Microsoft coco: Common objects in context.
\newblock In {\em ECCV}, 2014.

\bibitem{Liu-2013}
Y.~Liu, J.~Gall, C.~Stoll, Q.~Dai, H.-P. Seidel, and C.~Theobalt.
\newblock Markerless motion capture of multiple characters using multiview
  image segmentation.
\newblock {\em TPAMI}, 2013.

\bibitem{Loper2015}
M.~Loper, N.~Mahmood, J.~Romero, G.~Pons-Moll, and M.~J. Black.
\newblock Smpl: A skinned multi-person linear model.
\newblock In {\em TOG}, 2015.

\bibitem{orinet2018}
C.~Luo, X.~Chu, and A.~Yuille.
\newblock Orinet: A fully convolutional network for 3d human pose estimation.
\newblock In {\em BMVC}, 2018.

\bibitem{luvizon20182d}
D.~C. Luvizon, D.~Picard, and H.~Tabia.
\newblock 2d/3d pose estimation and action recognition using multitask deep
  learning.
\newblock In {\em CVPR}, 2018.

\bibitem{martinez2017simple}
J.~Martinez, R.~Hossain, J.~Romero, and J.~J. Little.
\newblock A simple yet effective baseline for 3d human pose estimation.
\newblock In {\em ICCV}, 2017.

\bibitem{singleshotmultiperson2018}
D.~Mehta, O.~Sotnychenko, F.~Mueller, W.~Xu, S.~Sridhar, G.~Pons-Moll, and
  C.~Theobalt.
\newblock Single-shot multi-person 3d pose estimation from monocular rgb.
\newblock In {\em 3DV}, 2018.

\bibitem{Mehta2017}
D.~Mehta, S.~Sridhar, O.~Sotnychenko, H.~Rhodin, M.~Shafiei, H.-P. Seidel,
  W.~Xu, D.~Casas, and C.~Theobalt.
\newblock Vnect: Real-time 3d human pose estimation with a single rgb camera.
\newblock {\em TOG}, 2017.

\bibitem{Moreno-noguer2017}
F.~Moreno-noguer.
\newblock {3D Human Pose Estimation from a Single Image via Distance Matrix
  Regression}.
\newblock In {\em CVPR}, 2017.

\bibitem{mueller2018ganerated}
F.~Mueller, F.~Bernard, O.~Sotnychenko, D.~Mehta, S.~Sridhar, D.~Casas, and
  C.~Theobalt.
\newblock Ganerated hands for real-time 3d hand tracking from monocular rgb.
\newblock In {\em CVPR}, 2018.

\bibitem{Newell-16}
A.~Newell, K.~Yang, and J.~Deng.
\newblock Stacked hourglass networks for human pose estimation.
\newblock In {\em ECCV}, 2016.

\bibitem{Nie2017}
B.~X. Nie, P.~Wei, and S.-C. Zhu.
\newblock Monocular 3d human pose estimation by predicting depth on joints.
\newblock In {\em ICCV}, 2017.

\bibitem{Oikonomidis-12}
I.~Oikonomidis, N.~Kyriazis, and A.~A. Argyros.
\newblock Tracking the articulated motion of two strongly interacting hands.
\newblock In {\em CVPR}, 2012.

\bibitem{pavlakos2018ordinal}
G.~Pavlakos, X.~Zhou, and K.~Daniilidis.
\newblock Ordinal depth supervision for 3{D} human pose estimation.
\newblock In {\em CVPR}, 2018.

\bibitem{pavlakos2017coarse}
G.~Pavlakos, X.~Zhou, K.~G. Derpanis, and K.~Daniilidis.
\newblock Coarse-to-fine volumetric prediction for single-image 3d human pose.
\newblock In {\em CVPR}, 2017.

\bibitem{pons2015dyna}
G.~Pons-Moll, J.~Romero, N.~Mahmood, and M.~J. Black.
\newblock Dyna: A model of dynamic human shape in motion.
\newblock {\em TOG}, 2015.

\bibitem{Ramakrishna2012}
V.~Ramakrishna, T.~Kanade, and Y.~Sheikh.
\newblock Reconstructing 3d human pose from 2d image landmarks.
\newblock In {\em CVPR}, 2012.

\bibitem{Rogez2016}
G.~Rogez and C.~Schmid.
\newblock {MoCap-guided Data Augmentation for 3D Pose Estimation in the Wild}.
\newblock In {\em NIPS}, 2016.

\bibitem{romero2017embodied}
J.~Romero, D.~Tzionas, and M.~J. Black.
\newblock Embodied hands: Modeling and capturing hands and bodies together.
\newblock {\em TOG}, 2017.

\bibitem{relativeposeBMVC18}
M.~R. Ronchi, O.~Mac~Aodha, R.~Eng, and P.~Perona.
\newblock It's all relative: Monocular 3d human pose estimation from weakly
  supervised data.
\newblock In {\em BMVC}, 2018.

\bibitem{Sharp-15}
T.~Sharp, C.~Keskin, D.~Robertson, J.~Taylor, J.~Shotton, D.~Kim, C.~Rhemann,
  I.~Leichter, A.~Vinnikov, Y.~Wei, et~al.
\newblock Accurate, robust, and flexible real-time hand tracking.
\newblock In {\em CHI}, 2015.

\bibitem{Shotton2011}
J.~Shotton, A.~Fitzgibbon, M.~Cook, and T.~Sharp.
\newblock {Real-time human pose recognition in parts from single depth images}.
\newblock In {\em CVPR}, 2011.

\bibitem{simon2017hand}
T.~Simon, H.~Joo, I.~Matthews, and Y.~Sheikh.
\newblock Hand keypoint detection in single images using multiview
  bootstrapping.
\newblock In {\em CVPR}, 2017.

\bibitem{Sridha-15}
S.~Sridhar, F.~Mueller, A.~Oulasvirta, and C.~Theobalt.
\newblock Fast and robust hand tracking using detection-guided optimization.
\newblock In {\em CVPR}, 2015.

\bibitem{sridhar2016real}
S.~Sridhar, F.~Mueller, M.~Zollh{\"o}fer, D.~Casas, A.~Oulasvirta, and
  C.~Theobalt.
\newblock Real-time joint tracking of a hand manipulating an object from rgb-d
  input.
\newblock In {\em ECCV}, 2016.

\bibitem{Sridhar-13}
S.~Sridhar, A.~Oulasvirta, and C.~Theobalt.
\newblock Interactive markerless articulated hand motion tracking using {RGB}
  and depth data.
\newblock In {\em ICCV}, 2013.

\bibitem{Sun2017}
X.~Sun, J.~Shang, S.~Liang, and Y.~Wei.
\newblock {Compositional Human Pose Regression}.
\newblock In {\em ICCV}, 2017.

\bibitem{Sun_2018_ECCV}
X.~Sun, B.~Xiao, F.~Wei, S.~Liang, and Y.~Wei.
\newblock Integral human pose regression.
\newblock In {\em ECCV}, 2018.

\bibitem{Taylor2000}
C.~J. Taylor.
\newblock Reconstruction of articulated objects from point correspondences in a
  single uncalibrated image.
\newblock {\em CVIU}, 2000.

\bibitem{Tekin2016}
B.~Tekin, A.~Rozantsev, V.~Lepetit, and P.~Fua.
\newblock {Direct Prediction of 3D Body Poses from Motion Compensated
  Sequences}.
\newblock In {\em CVPR}, 2016.

\bibitem{tompson2014joint}
J.~J. Tompson, A.~Jain, Y.~LeCun, and C.~Bregler.
\newblock Joint training of a convolutional network and a graphical model for
  human pose estimation.
\newblock In {\em NIPS}, 2014.

\bibitem{toshev2014deeppose}
A.~Toshev and C.~Szegedy.
\newblock Deeppose: Human pose estimation via deep neural networks.
\newblock In {\em CVPR}, 2014.

\bibitem{Tzionas-16}
D.~Tzionas, L.~Ballan, A.~Srikantha, P.~Aponte, M.~Pollefeys, and J.~Gall.
\newblock Capturing hands in action using discriminative salient points and
  physics simulation.
\newblock {\em IJCV}, 2016.

\bibitem{varol2018bodynet}
G.~Varol, D.~Ceylan, B.~Russell, J.~Yang, E.~Yumer, I.~Laptev, and C.~Schmid.
\newblock Bodynet: Volumetric inference of 3d human body shapes.
\newblock In {\em ECCV}, 2018.

\bibitem{ijcai2018-136}
M.~Wang, X.~Chen, W.~Liu, C.~Qian, L.~Lin, and L.~Ma.
\newblock Drpose3d: Depth ranking in 3d human pose estimation.
\newblock In {\em IJCAI}, 2018.

\bibitem{Wei2016}
S.-E. Wei, V.~Ramakrishna, T.~Kanade, and Y.~Sheikh.
\newblock Convolutional pose machines.
\newblock In {\em CVPR}, 2016.

\bibitem{yang20183d}
W.~Yang, W.~Ouyang, X.~Wang, J.~Ren, H.~Li, and X.~Wang.
\newblock 3d human pose estimation in the wild by adversarial learning.
\newblock In {\em CVPR}, 2018.

\bibitem{Ye-16}
Q.~Ye, S.~Yuan, and T.-K. Kim.
\newblock Spatial attention deep net with partial pso for hierarchical hybrid
  hand pose estimation.
\newblock In {\em ECCV}, 2016.

\bibitem{zhang20163d}
J.~Zhang, J.~Jiao, M.~Chen, L.~Qu, X.~Xu, and Q.~Yang.
\newblock 3d hand pose tracking and estimation using stereo matching.
\newblock {\em arXiv preprint arXiv:1610.07214}, 2016.

\bibitem{zhou2017towards}
X.~Zhou, Q.~Huang, X.~Sun, X.~Xue, and Y.~Wei.
\newblock Towards 3d human pose estimation in the wild: a weakly-supervised
  approach.
\newblock In {\em ICCV}, 2017.

\bibitem{zimmermann2017learning}
C.~Zimmermann and T.~Brox.
\newblock Learning to estimate 3d hand pose from single rgb images.
\newblock In {\em ICCV}, 2017.

\end{thebibliography}
